\documentclass[10pt,twocolumn,letterpaper]{article}

\usepackage{cvpr}              % To produce the CAMERA-READY version
\usepackage{svg}

\usepackage[dvipsnames]{xcolor}

\usepackage{graphicx}
\usepackage{amsmath}
\usepackage{amssymb}
\usepackage{booktabs}
\usepackage{algorithm}
\usepackage{algpseudocode}

\definecolor{cvprblue}{rgb}{0.21,0.49,0.74}
\usepackage[pagebackref,breaklinks,colorlinks,citecolor=cvprblue]{hyperref}

\def\paperID{14462} % *** Enter the Paper ID here
\def\confName{CVPR}
\def\confYear{2024}

\title{Dual Band Thermal Videography:\\Separating Time-Varying Reflection and Emission Near Ambient Conditions}

\author{
Sriram Narayanan \quad
Mani Ramanagopal \quad
Srinivasa G.\ Narasimhan \\[3mm]
\textnormal{Carnegie Mellon University} \\[1mm]
\href{https://dual-band-thermal.github.io/}{\texttt{dual-band-thermal.github.io}}
}

\begin{document}
\maketitle
\begin{abstract}
% Long-wave infrared radiation captured by a thermal camera consists of two components: (a) light from the environment reflected or transmitted by a surface, and (b) light emitted by the surface after undergoing heat transport through the object and exchanging heat with the surrounding environment. Separating these components is essential for understanding object properties such as emissivity, temperature, reflectance and shape. Previous thermography studies often assume that only one component is dominant (e.g., in welding) or that the second component is constant and can be subtracted. However, in near-ambient conditions, which are most relevant to computer vision applications, both components are typically comparable in magnitude and vary over time. We introduce the first method that separates reflected and emitted components of light in videos captured by two thermal cameras with different spectral sensitivities. We derive a dual-band thermal image formation model and develop algorithms to estimate the surface's emissivity and its time-varying temperature while isolating a dynamic background. We quantitatively evaluate our approach using carefully calibrated emissivities for a range of materials and show qualitative results on complex everyday scenes, such as a glass filled with hot liquid and people moving in the background. 
Long-wave infrared radiation captured by a thermal camera includes (a) emission from an object governed by its temperature and emissivity, and (b) reflected radiation from the surrounding environment. Separating these components is a long-standing challenge in thermography. Even when using multiple bands, the problem is under-determined without priors on emissivity. This difficulty is amplified in near ambient conditions, where emitted and reflected signals are of comparable magnitude. We present a dual-band thermal videography framework that reduces this ambiguity by combining two complementary ideas at a per-pixel level: (i) spectral cues (ratio of emissivity between bands is unknown but fixed), and (ii) temporal cues (object radiation changes smoothly while background radiation changes rapidly). We derive an image formation model and an algorithm to jointly estimate the object's emissivity at each band, and the time-varying object and background temperatures. Experiments with calibrated and uncalibrated emissivities in everyday scenes (e.g., coffee pot heating up, palm print on mirrors, reflections of moving people), demonstrate robust separation and recovery of temperature fields. 
% We will release code and data upon acceptance.
%and , which we show is essential for accurately estimating an object's material and shape.
\end{abstract}    
\vspace{-0.1in}
\section{Introduction}
\label{sec:intro}

Long-Wave Infrared (LWIR, 8-14\, \textmu m) radiation received by a thermal camera consists of two components: (1) the reflection or transmission of radiation originating from the surrounding environment, akin to traditional light transport, and (2) the object’s own thermal emission, governed by internal heat transport and heat exchange with its surroundings. 
%Unlike visible light, where appearance is largely determined by illumination and surface reflectance or transmittance, thermal appearance depends on the object’s temperature, emissivity, and the thermal radiation and characteristics of the environment. 
Accurately disentangling these factors  is crucial to correctly interpret thermal imagery for a range of scientific and engineering applications \cite{araujo2017multi, vollmer2020infrared}.

%\cite{physicsOfThermalRad, SUZUKI201919, Ogasawara2017, Grimming2023, Zhang:24-1, Zhang:23} 

In general, separating the reflected and emitted components of thermal radiation is a highly under-constrained problem~\cite{jeanclaudePyro}. As a result, most prior approaches assume that one of the components is negligible or known \emph{a~priori} \cite{Chrzanowski2001NONCONTACTTM,ZHANG20091} . For example, in industrial scenarios like welding and defect inspection, objects are significantly hotter than their surroundings, allowing background component to be neglected \cite{spotwelding, ActiveThermoReflectometry}. Other studies employ controlled environments where the background component is spatially uniform or measured ahead of time \cite{Ogasawara2017} or inferred from polarization~\cite{SUZUKI201919}. Some methods assume stationary backgrounds \cite{TanakaFar,Narayanan2024Shape} or steady-state object emissions that can be subtracted out \cite{thermalNLOS}. 
%Others depend on prior knowledge of emissivity and ambient temperature \cite{}. 
However, most scenes of relevance to computer vision are near-ambient conditions where the reflected and emitted components are comparable in magnitude, time-varying, and spatially complex, making such assumptions unreliable or impractical.

In this work, we present a novel approach to disentangle the time-varying reflected and emitted radiation components using thermal videos captured within two sub-bands of LWIR. Our method builds on two key insights: (1) an object’s spectral emissivity can vary with wavelength even within LWIR - contrary to the conventional ``gray-body'' assumption \cite{araujo2017multi} and (2) heat transport within objects evolves smoothly over time, while background changes abruptly. 

We derive the dual-band image formation model as a linear combination of the reflection and emission components. Given thermal videos captured with two spectral bandpass filters, we consider both calibrated and uncalibrated scenarios.  In the calibrated setting, we estimate material emissivity \emph{a priori} from contact measurements of temperature using thermocouples where the background is a blackbody at known temperature. We then use the calibrated emissivity to separate reflection and emission in new scenes with those materials. To address the ill-posed uncalibrated case, our optimization exploits the differences in temporal variations in reflection vs. emission using a novel reconstruction loss to jointly estimate object emissivity and time varying object temperature and background radiation.

We validate our calibrated and uncalibrated approaches through both simulation and real experiments. We show that our calibrated emissivities of materials match those in the ECOSTRESS spectral library \cite{ECOSTRESSv1, ECOSTRESSv2}. We also show that our results match the reflector approach recommended by FLIR Systems \cite{flir_thermography_reference_2019} for higher emissivity materials and improve on that approach for lower emissivity materials. Using simulations, we compare our uncalibrated method against existing baselines \cite{Grimming2023, araujo2017multi, dualWavelengthPryo} across diverse materials  under varying noise conditions. 

%Experimentally, we design a novel calibration setup that simultaneously records object and reflected temperatures with thermocouples, enabling empirical emissivity verification across multiple materials.
Finally, we demonstrate the separation of dynamic emitted and reflected thermal components in real-world videos. Using dual-band thermal imaging, we recover emission and reflection videos in challenging scenes. The slow emission due to heat transport from a coffee pot containing a hot liquid is separated from reflections of people moving near it. A glass plate heated by a hot air gun increases its emission and is separated from a reflection of text written on a surface with cold and wet fingers. A palm print dissipating on a cold bulb is separated from higher reflection intensities due to a person. Please watch the supplementary video for the best visualizations. Our results show that thermal imaging, when analyzed spectrally and temporally, can reveal rich scene information inaccessible to  visible-light cameras.

%We validate our method in simulation and compare against baselines~\cite{Grimming2023, araujo2017multi} across materials with different emissivities sourced from the ECOSTRESS spectral library~\cite{ECOSTRESSv1, ECOSTRESSv2}, under a range of thermal camera noise levels. We then experimentally measure and verify emissivity for diverse objects using a new calibration setup that jointly captures the object and its reflection while recording their temperatures with thermocouples.

%Finally, we visually demonstrate the separation of object emission from reflected background for the first time in dynamic scenes, in both calibrated and uncalibrated conditions. Using dual-band thermal video capture, we show results on challenging real-world cases such as a glass containing hot liquid with people moving behind it, and distinguishing a palm print on a surface from its reflection. As such, this work represents an important step in utilizing thermal imaging to capture new scene information beyond the capability of visible imaging.

\section{Related Work}
\label{sec:relatedwork}

\noindent{\bf{Point measurements:}}
Separating reflected and emitted radiation by an object has a long history \cite{physicsOfThermalRad, SUZUKI201919, Ogasawara2017, Grimming2023, Zhang:24-1, Zhang:23} but with no universally accepted technique applicable in general situations \cite{araujo2017multi, jeanclaudePyro}. The survey in \cite{araujo2017multi} concludes that the best works are still the early ones performing dual \cite{dualWavelengthPryo} and multi-band \cite{Saunders_2003} measurements. However, these methods make a ``gray body'' assumption and focus on measurements at a single point and do not demonstrate spatially varying signals of complex scenes relevant to computer vision. Our method outperforms such approaches.

\noindent{\bf{Pyrometry:}}
Our work differs from studies in pyrometry \cite{PyrometryNicodemus:65,araujo2017multi}, remote sensing \cite{RemoteSensSCHMUGGE2002189}, and related non-contact thermometry \cite{Chrzanowski2001NONCONTACTTM,ZHANG20091} in several aspects. These methods often assume single, dual or multi-band measurements with either known/parameterized emissivity or conditions where emission dominates. Many works focus on hotter targets (200–2000\,°C) and MWIR sensing where reflectivity is lower \cite{araujo2017multi,ActiveThermoReflectometry,vollmer2020infrared,spotwelding}. We instead operate near ambient temperatures with LWIR (8–14\,\textmu m) sensors, where reflectance is higher and background radiation strongly contaminates measurements. Our method explicitly models and compensates for temporally varying background radiation.

\noindent{\bf{Reducing reflections and enhancing contrast:}}
Recent approaches reduce reflections using polarization \cite{SUZUKI201919} or enhance image contrast using priors \cite{Ogasawara2017,Grimming2023,BCP-1,BCP-2,bao2023heat}.  For e.g., Blackbody Channel Prior (BCP) \cite{Grimming2023} boosts contrast assuming locally brightest pixels approximate blackbody emitters. However, BCP’s assumption breaks for cold objects near ambient temperatures and where both object and background vary over time. In contrast, our work is focused on physically-based separation of reflection and emission components that are both dynamic near ambient conditions. We demonstrate that our approach performs better than BCP.

% Our work also differs from prior studies in pyrometry \cite{PyrometryNicodemus:65, araujo2017multi}, remote sensing \cite{RemoteSensSCHMUGGE2002189}, and other related fields \cite{Chrzanowski2001NONCONTACTTM, ZHANG20091} that estimate emitted light using single, dual, or multi-band thermal measurements in several key aspects: (i) we account for temporal variations in background radiation and its influence, (ii) our focus is on scenes operating near ambient temperatures, characterized by low signal-to-noise ratios, and (iii) we utilize LWIR cameras operating in the 8–14 µm range, where objects exhibit higher reflectivity and background radiation significantly impacts measurements. In contrast, \cite{araujo2017multi, ActiveThermoReflectometry} show that many works target object temperatures ranging from 200°C upto 2000°C, where emitted radiation dominates in the mid-wave infrared spectrum and materials have lower reflectivity \cite{vollmer2020infrared}. 

%\vspace{-0.4cm}
\noindent{\bf{Physics Based Vision with Thermal:}}
The use of thermal cameras for vision tasks is gaining attention, particularly those exploiting the dominant radiative components, either reflection \cite{thermalNLOS, Kaga2019} or emission \cite{Dashpute_2023_CVPR, SFTR, TanakaFar,  Narayanan2024Shape, Ramanagopal_2024_CVPR} enabling novel ways to solve many vision tasks. In particular, Dashpute et al. \cite{Dashpute_2023_CVPR} recover properties such as emissivity and diffusivity but the method requires access to surface temperatures and ignores the influence of background. Similarly, many works use the emitted or reflected light to recover material or geometric properties such as shape \cite{Narayanan2024Shape, SFTR, KitazawaICCP25,Kushida_2025_ICCV}, albedo \cite{Ramanagopal_2024_CVPR, yuan2025ordinalityvisiblethermalimageintensities}, pose \cite{thermalNLOS} and trajectory \cite{SheininCVPR24} but assume a thermally static background or known object temperatures and hence can subtract the first frame to ignore the effect of reflections \cite{TanakaFar} or ambient emissions. \cite{11143836} train a neural network to estimate transparent shape from computed degree of polarization in the presence of both reflection and emission. But they avoid reflections of backgrounds (e.g. people) similar in magnitude to the object temperatures.
% to obtain object temperature.

%\vspace{-0.4cm}
\noindent{\bf{Light Transport Decomposition:}} Our research is similar in spirit to many physically based works that separate different light transport components including diffuse vs. specular \cite{Lin2002diffspecular, Dave:22, 10030760, 8614311}, reflection vs. transmission \cite{RefTransPatrick,RefTransLevin, Shih_2015_CVPR, RefTransXue}, reflectance vs. shading \cite{Shafer1985}, direct vs. global \cite{Nayar2006, Boxin2024}, and more \cite{Sun:22}. But we operate in thermal domain where both heat and light transports are involved. Tanaka et al. \cite{TanakaFar} decompose the time varying heat transport emissions into ad-hoc diffuse and global components with a constant background reflection that can be subtracted out while our physically-grounded approach estimates the emissivity and temperatures of the object and background.

\section{Thermal Image Formation Model}
%We briefly highlight components of thermal radiation that are crucial to the image formation process of a thermal camera. 
Thermal cameras convert the incoming radiation in the Long Wave Infrared (LWIR) spectrum to digital images. 
The radiation emitted by a blackbody at temperature $T$ and wavelength $\lambda$ follows Planck’s spectral distribution \cite{physicsOfThermalRad}:
\begin{equation}
    L_b (\lambda, T_{bb}) = \frac{c_1}{\lambda^5 \exp\left({\frac{c_2}{\lambda T_{bb}}}\right) - 1},
    \label{eq:plancks}
\end{equation}
where, $c_1, c_2$ are known radiation constants. 
However, real-world objects emit only a fraction of this radiation, determined by their spectral emissivity $\epsilon (\lambda)$, where $0 \leq \epsilon(\lambda) \leq 1$. Thus, the emitted radiation from a real object is:
\begin{equation}
    L(\lambda, T) = \epsilon(\lambda)L_b(\lambda, T)
\end{equation}
A thermal camera uses this emitted radiation to infer the temperature of an object, therefore it is crucial to account for the object's emissivity to obtain accurate measurements. But, the image formation model of a thermal camera is more involved than directly measuring the emitted radiation.

\begin{figure}[t]
    \centering
    \includegraphics[width=0.9\linewidth, trim={0 14cm 5cm 0}, clip]{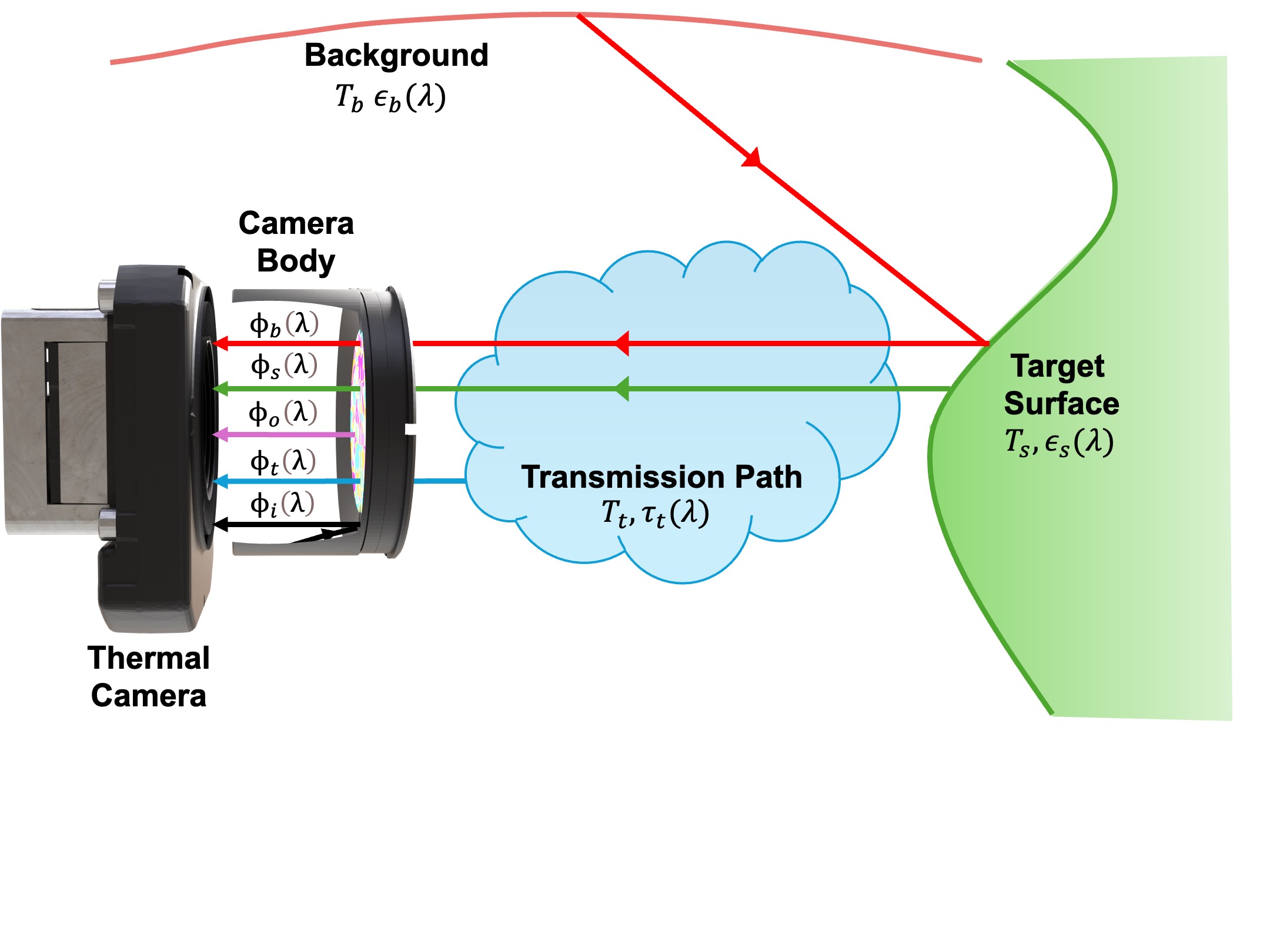}
    \caption{Image formation in a thermal camera comprises of radiation from the object $\Phi_s$, reflection $\Phi_b$, optics $\Phi_o$, transmission path $\Phi_t$ and its internal components $\Phi_i$.}
    \label{fig:thermal_cam_formation}
    \vspace{-0.2in}
\end{figure}

Figure \ref{fig:thermal_cam_formation} shows various components of thermal radiation that contribute to the image formation. The thermal radiation at wavelength $\lambda$ comprises of five main components: radiation $\Phi_s(\lambda)$ emitted by the target surface, radiation $\Phi_b(\lambda)$ emitted by the background objects and reflected or transmitted through the target, radiation  $\Phi_t(\lambda)$ emitted by the transmission path, radiation $\Phi_o(\lambda)$ emitted by the optics and radiation $\Phi_i(\lambda)$ emitted by the internal components of the thermal camera. The total radiation received at the camera is a weighted sum of these components, where the weights depend on the emissivity $\epsilon(\lambda)$, transmissivity $\tau(\lambda)$ and reflectivity $r(\lambda)$. From Kirchoff's law:
% Krichoff's law states that $\epsilon(\lambda) \cite{physicsOfThermalRad}
\begin{equation}
    \epsilon(\lambda) + \tau(\lambda) + r(\lambda) = 1
\end{equation}
Then, we can write the total spectral radiation leaving the object's surface as a weighted combination of an object's emitted, reflected or transmitted radiation:
\begin{equation}
    \Phi_A(\lambda) = \epsilon_s(\lambda)\Phi_s(\lambda) + (1 - \epsilon_s(\lambda))\Phi_b(\lambda) 
    \label{phi_A_eq}
\end{equation}
Similarly, the radiation reaching the optics is a combination of attenuated radiation from the object's surface and radiation emitted by the transmission path,
\begin{equation}
    \Phi_B(\lambda) = \tau_t(\lambda)\big( \Phi_A(\lambda) \big) + (1 - \tau_t(\lambda))\Phi_t (\lambda)
\end{equation}
% \begin{equation}
%     \tau_t\big(\epsilon_s\Phi_s + (1 - \epsilon_s)\Phi_b \big) + (1 - \tau_t)\Phi_t 
% \end{equation}
%For short scene distances, transmissivity $\tau \approx 1$. However, additional attenuation and reflection occur due to the camera optics and filters. 
A portion of the camera's own thermal radiation is also reflected by the optical elements, an effect known as the narcissus effect \cite{SFTR}. Accounting for this, the total radiation reaching the detector at wavelength $\lambda$ is:
% Typically, we assume the transmission path to have negligible reflectivity and for small scene distances, the transmissivity $\tau_t \approx 1$. Further, camera optics and optical filters also reflect and attenuate infrared light. This effect of thermal light from camera body's reflected off from the optical components is called as narcissus effect \cite{SFTR}. Therefore, the total thermal radiation reaching the detector at a particular wavelength $\lambda$ accounting for the narcissus effect can be written as follows,
\begin{equation}
    \Phi_{C} (\lambda) =  \tau_o(\lambda) \Phi_B(\lambda) + \epsilon_o(\lambda) \Phi_o(\lambda) + r_o(\lambda) \Phi_i(\lambda)
    \label{eq:total_flux_in}
\end{equation}
here, $\epsilon_o, \tau_o$ and $r_o$ denote the emissivity, transmissivity and reflectivity of the optical components respectively. The total incoming radiation $\Phi_{tot}$ at the detector is obtained by integrating across the camera’s sensitivity spectrum
% Here, we drop the dependence of $\lambda$ on the right for simplicity. $\tau_o$ and $r_o$ are the transmissivity and reflectivity of the optical component. 
% Integrating the total input radiation across the camera's sensitivity spectrum $\lambda_{\min} \to \lambda_{\max}$ yields the total incoming radiation $\Phi_{tot}$ to the detector. 
\begin{equation}
    \Phi_{tot} = \int_{\lambda_{\min}}^{\lambda_{\max}} \Phi_{C}(\lambda)\mathrm{d}\lambda
\end{equation}
\vspace{-0.05in}
% The thermal camera signal $W$ is proportional to this incoming radiation. 
% Typically, the wavelength dependence on emissivity, transmissivity and reflectivity is dropped, but accounting for these components is crucial towards correctly disentangling these components.
% Thermal cameras provide a signal $U$ proportional to the total incoming radiation within its sensitivity spectrum. Radiation emitted by a blackbody within the sensitivity spectrum depends on its temperature and can be obtained by integral of Planck's law \ref{eq:plancks}. 
% From this, we can interpret any radiation $\Phi$ as a radiation emitted by a blackbody at an arbitrary temperature $T$. Re-writing Eq. \ref{eq:total_flux_in} in terms of thermal camera signal we get,
% From this, the incoming radiation $\Phi_{tot}$ can be interpreted as the radiation emitted by a blackbody at some arbitrary temperature $T_{bb}$. We can then write Eq. \ref{eq:total_flux_in} as a linear combination of thermal camera signals for some object emitting that radiation at some arbitrary temperature.
% \begin{equation}
%     U(T_{bb}) = \tau_o U(T_B) + \epsilon_o U(T_o) + r_o U(T_i)
% \end{equation}
% where, $T_B$

We make the following assumptions based on our imaging setup: (i) the camera-to-scene distance is small enough to neglect atmospheric attenuation ($\tau_t \approx 1$), (ii) radiation from the optics and internal camera elements remains constant across experiments, and (iii) we consider averaged values of $\epsilon, \tau$, and $r$ within a given sensitivity spectrum. Under these conditions, the total incoming radiation is:
\begin{equation} \Phi_{tot} = g \Phi_A + f \end{equation}
where $g = \tau_o$ and $f = \epsilon_o \Phi_o + r_o \Phi_i$ are the gain and offset corrections for the narcissus effect introduced by optical filters and internal camera elements. Unlike \cite{SFTR}, which corrects only for the offset by subtracting a reference frame, we apply both gain and offset corrections for better accuracy.

From Planck’s law (Eq. \ref{eq:plancks}), any radiation can be interpreted as blackbody radiation at some temperature $T$. After correcting for gain and offset, we express the thermal camera signal $U(T_{bb})$ for incoming radiation $\Phi_A$ (Eq. \ref{phi_A_eq}) as:
\begin{equation} 
U(T_{bb}) = \epsilon_s U(T_{o}) + (1-\epsilon_s) U(T_{b}) \label{eq:LT_ut} 
\end{equation}
where, $T_o$ is the object temperature, $T_b$ denotes an effective background temperature and
$U(T_{bb})$ relates pixel intensity to blackbody temperatures and is approximated using the Sakuma-Hattori model:
\begin{equation} 
U(T_{bb}) = \frac{c_1}{\exp{(\frac{c_2}{c_3T_{bb}+c_4})} - 1}, 
\end{equation}
with calibration constants $c_1, c_2, c_3$, and $c_4$ determined through curve fitting \cite{sakuma1982establishing}. Given that the Sakuma-Hattori function is nearly linear for small temperature differences around ambient conditions, \ie $U(T) = aT + b$. The calibration constants $a$ and $b$ are obtained using a blackbody at a known temperature (see supplementary).

\vspace{-0.4cm}
\paragraph{Band Limited Thermal Videos:} 
We capture $M$ multi-spectral thermal bands that lie within the camera's sensitivity spectrum of 8-14\, \textmu m. The captured pixel intensity within each band $m$ across an entire video is expressed as:
\begin{equation}
    I_m(t) = \epsilon_m U_m({T_{o}}(t)) + (1-\epsilon_m) U_m({T_{b}}(t)).
    \label{eq:multi_band_LT}
\end{equation}
As the background is composed of various objects with different emissivities and temperatures, we assume their effective contributions average out such that $T_b$ is independent of wavelength. 
For simplicity, we denote the measured pixel intensity $U(T_{bb})$ in Eq. \ref{eq:LT_ut} as $I$ and band emissivity as $\epsilon_m$. 
Function $U_m(T)$ that converts blackbody temperature to pixel intensity varies across bands $U_m(T) = a_m T + b_m$ due to varying optical properties and camera response function within the sensitivity spectrum.
Further, $t$ denotes the frame of the video indicating a time varying object temperature $T_{o}(t)$ and effective background temperature $T_{b}(t)$. 

\section{Method}
%In this section, we develop a novel theory of dual-band radiation thermography that utilizes multi-spectral thermal measurements to decompose incoming radiation into its constituent components. The incoming radiation $\Phi_{tot}$ can be categorized into several components: reflected light, emitted light, and light transmitted through the material. While reflected/transmitted light arises solely from light transport phenomena, emitted light is due to heat transport within and around the object. Emitted light serves as a key intermediary in the heat transport process, making it a vital element in thermographic analysis \cite{SFTR, Narayanan2024Shape, Srinivasan_Ramanagopal_2020, Dashpute_2023_CVPR}. 

%\subsection{Problem Definition}
In this section, we analyze thermal videos captured in two different spectral bands, where emissivity varies between them. Given thermal video data over $N$ timesteps, represented as $ \{I_m^{1},...,I_m^N\} $ for each band $ m \in \{1,2\} $, our goal is to decompose the thermal radiation into: the band emissivity \( \epsilon_m \), the time-varying object's temperature \( T_{o}(t) \), and the effective background temperature \( T_{b}(t) \) that governs the image formation process described in Eq. \ref{eq:multi_band_LT}.

\subsection{Calibrated Case}
A priori, we calibrate the spectral emissivity of the materials of interest. Then, we use the calculated emissivity to separate the dynamic reflected background and the emission in a new scene with the same materials. To calibrate, we use a thermocouple and measure temperature at a point on the material. Also, we place a blackbody at known temperature near the material to serve as background. We then capture thermal images of the material with different spectral filters (see Fig. \ref{fig:emiss-calib}). Here, the object and background temperatures are known using the thermocouple and the blackbody. Then, emissivity $\epsilon_m$ is computed as:

%To perform calibrated light decomposition, we use a reference blackbody emitter at a known temperature and measure the object's temperature with a thermocouple. This allows us to determine spectral emissivity across different bands using Eq. \ref{eq:multi_band_LT}. Given the known object and background temperatures, the emissivity $\epsilon_m$ is computed as:
% For calibrated light decomposition, we use a reference blackbody emitter at a known temperature and measure the object's temperature with a thermocouple to determine spectral emissivity across different bands using Eq. \ref{eq:multi_band_LT}. Given known object and background temperatures, the emissivity $\epsilon_m$ is computed as:
% In the calibrated light decomposition case, we use a reference blackbody emitter at known temperature and measure object's temperature using a thermocouple to identify the spectral emissivity of various bands based on Eq. \ref{eq:multi_band_LT}. With known object and background temperatures, the emissivity $\epsilon_m$ is given by,
\begin{equation}
    \epsilon_m = \dfrac{I_m - U_m(T_{b})}{U_m(T_{o}) - U_m(T_{b})}
    \label{eq:emiss_calib_eq}
\end{equation}

\vspace{-0.4cm}
\paragraph{Estimating $T_{o}$ and $T_{b}$ from Known Emissivities:}
The emissivities $\epsilon_m$ for two spectral bands ($m \in \{1,2\}$) obtained from calibration are then used for a new scene with unknown object and background temperature. The object's time-varying temperature $T_o$ and the background temperature $T_b$ are estimated using closed-form expressions. Expanding Eq. \ref{eq:multi_band_LT} for two bands ($m \in \{1,2\}$), we have:
% Given the emissivity $\epsilon_m$ of the two bands $\in [1, 2]$ from the calibration procedure we can estimate object and ambient reflected temperature as closed from expressions. The expanded version of Eq. \ref{eq:multi_band_LT} for the two bands can be written as:
% The pixel intensities $I_i$ and $I_j$ for any two bands $i$ and $j$ can be written of the form:
\begin{equation}
    \begin{aligned}
        I_1 &= \epsilon_1 a_1 (T_{o} - T_{b}) + a_1 T_{b} + b_1 \\
        I_2 &= \epsilon_2 a_2 (T_{o} - T_{b}) + a_2 T_{b} + b_2\,.
    \end{aligned}
    \label{eq:2bands_LT_known}
\end{equation}
Solving for $T_{\text{o}}$ and $T_{\text{b}}$ form the above equation, we get:
\begin{equation}
    \begin{aligned}
        T_{o} &= \frac{a_1 (I_2 - b_2) (\epsilon_1 - 1) - a_2 (I_1 - b_1) (\epsilon_2 - 1)}{a_1 a_2 (\epsilon_1 - \epsilon_2)} \\
        T_{b} &= \frac{a_1 (I_2 - b_2) \epsilon_1 - a_2 (I_1 - b_1) \epsilon_2}{a_1 a_2 (\epsilon_1 - \epsilon_2)}\,.
    \end{aligned}
    \label{eq:solve_ToTb}
\end{equation}
% This formulation enables estimation of object and background temperatures based on known emissivities from the calibration process.

\subsection{Uncalibrated Case}

When object emissivities are unknown, jointly estimating object and background temperatures becomes ill-posed. Although the thermal light transport in Eq.~\ref{eq:LT_ut} is in principle linearly independent across bands with differing emissivities, the near-linear behavior of the Sakuma–Hattori function $U(T)$ close to ambient conditions collapses the effective rank of the system. As a result, the two-band image formation model can be rewritten compactly as:
\begin{equation}
\begin{bmatrix}
\dfrac{I_1 - b_1}{a_1} \\[10pt]
\dfrac{I_2 - b_2}{a_2}
\end{bmatrix}
=
\begin{bmatrix}
\epsilon_1 & 1-\epsilon_1 \\
\epsilon_2 & 1-\epsilon_2
\end{bmatrix}
\begin{bmatrix}
T_{\text{o}} \\[5pt]
T_{\text{b}}
\end{bmatrix},
\label{eq:img_form_matform_vfgbg}
\end{equation}
where the constants $a_i$ and $b_i$ are absorbed into the left-hand side. Writing this compactly as $\mathbf{I} = \mathbf{E}\mathbf{T}$ makes the structure apparent: the matrix $\mathbf{E}$ has rank 2. This follows from our reasonable assumption that $T_b$ is independent of wavelength. Extending this formulation to $M$ spectral bands and $N$ timesteps yields $\mathbf{I}\in\mathbb{R}^{M\times N}$, $\mathbf{E}\in\mathbb{R}^{M\times 2}$, and $\mathbf{T}\in\mathbb{R}^{2\times N}$, with $\mathbf{E}$ still of rank 2 by construction. Hence, additional constraints are required to recover $T_{\text{o}}$ and $T_{\text{b}}$ in the uncalibrated case.

% \subsection{Uncalibrated Thermal Light Decomposition}
% When the emissivity of the objects in the scene is unknown, estimating the object and ambient temperatures becomes an ill-posed problem. Although the thermal light transport described in Eq. \ref{eq:LT_ut} for various bands with different emissivities can be linearly independent, the linear behavior of the Sakuma-Hattori function $U(T)$ near ambient conditions reduces the overall rank of the system. Thus, we can rewrite the image formation across two spectra in matrix form as,
% \begin{equation}
%     \begin{bmatrix}
%         \dfrac{I_1 - b_1}{a_1} \\[10pt]
%         \dfrac{I_2 - b_2}{a_2}
%     \end{bmatrix}
%     =
%     \begin{bmatrix}
%         \epsilon_1 & 1 - \epsilon_1 \\
%         \epsilon_2 & 1 - \epsilon_2
%     \end{bmatrix}
%     \begin{bmatrix}
%         T_{\text{o}} \\[5pt]
%         T_{\text{b}}
%     \end{bmatrix}
%     \label{eq:img_form_matform_vfgbg}
% \end{equation}
% Here, the constants $a_i$ and $b_i$ can be pre-computed and included in the left-hand side of the equation. Writing the above system in the form $\mathbf{I} = \mathbf{E}\mathbf{T}$ reveals that the rank of $\mathbf{E}$ is 2. This formulation generalizes to $M$ spectral bands and $N$ timesteps while still maintaining a rank of 2.
% In this generalized case, the resulting matrices $\mathbf{I} \in \mathbb{R}^{M\times N}$, $\mathbf{E} \in \mathbb{R}^{M \times 2}$, and $\mathbf{T} \in \mathbb{R}^{2\times N}$ demonstrate that $\mathbf{E}$ remains of rank 2 by definition. 
% This shows additional constraints are necessary towards solving the above system.

\vspace{-0.1in}
\subsubsection{Thermography in Dynamic Scenes}
% Dynamic nature of the scene is much more of common phenomenon in thermal compared to visible. Even though an object could be in a static state of motion with visible light transport being constant, it might not be in thermal equilibria with the surrounding, leading to dynamic thermal appearance every instant. In this work, we aim leverage the dynamic nature of change in object temperature with respect to background motion to develop novel constraints for thermography.
% Dynamic nature of thermal scenes is much more common than a 
% \sriram{Motivation for constraints.}
% Even when an object is stationary and visible light transport remains constant, it may not be in thermal equilibrium with its surroundings, leading to continuous variation in thermal.
The appearance of an object is often more dynamic in the thermal spectrum than in the visible spectrum. Even when the object remains stationary and light transport is constant at millisecond timescales, it may not be in thermal equilibrium with its surroundings, leading to continuous variations in thermal emission over time. Fortunately, heat transport evolves gradually, while background reflections typically change abruptly and lack correlation with the object’s emission. Leveraging this observation, we introduce novel constraints that  disentangle reflection and emission.
%In this work, we leverage these dynamic temperature changes in relation to background motion to develop novel constraints for thermography.

%\vspace{-0.4cm}
\noindent{\bf Prior on object temperature $T_{\text{o}}$:}
The change in object temperature is governed by the heat transfer process that occurs within and around it governed by the standard heat transport equation \cite{vollmer2020infrared,Narayanan2024Shape,Dashpute_2023_CVPR}. Ramanagopal et al. \cite{Ramanagopal_2024_CVPR} show that with short capture timescales, conduction is negligible, simplifying the system to an ordinary differential equation (ODE).  The solution to this ODE follows an exponential form serving as a valuable prior for addressing the ill-posed nature of the system in Eq. \ref{eq:img_form_matform_vfgbg}.

\noindent{\bf Constraints from a static background:} In many practical scenarios, the change in background radiation observed by the camera occurs in smaller portions of the scene. This is especially true when the target surface is glossy/specular. In such cases, one can write a differential form of Eq. \ref{eq:multi_band_LT},
\begin{equation}
        \frac{1}{a_m}\frac{\partial I_m}{\partial t} = \epsilon_m\frac{\partial T_{o}}{\partial t} + (1 - \epsilon_m) \frac{\partial T_{b}}{\partial t}
        \label{eq:diffEq_nobgm_prev}
\end{equation}
% \begin{equation}
%         \frac{1}{a_1}\frac{\partial I_1}{\partial t} = \epsilon_1\frac{\partial T_{o}}{\partial t} + (1 - \epsilon_1) \frac{\partial T_{b}}{\partial t}
%         \label{eq:diffEq_nobg1_prev}
% \end{equation}
% \begin{equation}
%         \frac{1}{a_2}\frac{\partial I_2}{\partial t} = \epsilon_2\frac{\partial T_{o}}{\partial t} + (1 - \epsilon_2) \frac{\partial T_{b}}{\partial t}\\
%         \label{eq:diffEq_nobg2_prev}
% \end{equation}
At pixels where the change in background radiation is insignificant, the differential change is negligible. Taking the ratio of the dual bands in Eq. \ref{eq:diffEq_nobgm_prev} yields a constant ratio $k_1$:
% At pixels where the change in background radiation is insignificant, the differential change is negligible. Taking the ratio of Eq. \ref{eq:diffEq_nobg1_prev} to \ref{eq:diffEq_nobg2_prev} yields a constant ratio $k_1$:
% In pixel locations where the change in background radiation is insignificant, the above sets of equations reduce to,
% \begin{equation}
%         \frac{1}{a_1}\frac{\partial I_1}{\partial t} = \epsilon_1\frac{\partial T_{o}}{\partial t}
%         \label{eq:diffEq_nobg1}
% \end{equation}
% \begin{equation}
%         \frac{1}{a_2}\frac{\partial I_2}{\partial t} = \epsilon_2\frac{\partial T_{o}}{\partial t}
%         \label{eq:diffEq_nobg2}
% \end{equation}
% The ratio between the differential images from Eq. \ref{eq:diffEq_nobg1} and \ref{eq:diffEq_nobg2} yields a constant ratio $k_1$ across a thermal video, where,
\begin{equation}
    k_1 = \frac{\epsilon_2}{\epsilon_1}  = \frac{a_1 \, \dfrac{\partial I_2}{\partial t}}{a_2 \, \dfrac{\partial I_1}{\partial t}}
    \label{eq:k1_const}
\end{equation}

%\vspace{-0.3cm}
\noindent{\bf Constraints from a dynamic background:} 
From Eq \ref{eq:multi_band_LT} we can decompose the image for any spectral band as,
% \begin{equation}
%     I_m = \epsilon_m U(T_{o}) + (1-\epsilon_m)U(T_{b})
% \end{equation}
\begin{equation}
    I_m(t) = \epsilon_m U(T_{o}(t)) + (1-\epsilon_m)U(\bar{T}_{b} + T_{b}^*(t))
\end{equation}
where $\bar{T}_{b}$ is some mean background temperature and $T_{b}^*(t)$ represents the differential change in the background from the mean. This can be separated into:
\begin{multline}
    I_m(t) = \underbrace{\epsilon_m U(T_{o}(t)) + (1-\epsilon_m)U(\bar{T}_{b})}_{\tilde{I}_m (t)} \\+ (1 - \epsilon_m) a_m T_{b}^*(t)
\end{multline}
% here, $\tilde{I}_m (t)$ is an emission dominant smoothly varying signal from our prior based on object temperature. 
where $\tilde{I}_m (t)$ represents a smoothly varying, emission-dominant signal derived from the object's temperature prior.
Here we assume that object and background temperatures vary in an un-correlated fashion. 
From this, given smooth signals in two spectral bands, we can express:
\begin{equation}
    I_1(t) - \tilde{I}_1(t) = (1 - \epsilon_1) a_1 T_{b}^*(t)
    \label{eq:diffEq_nofg1}
\end{equation}
\begin{equation}
    I_2(t) - \tilde{I}_2(t) = (1 - \epsilon_2) a_2 T_{b}^*(t)
    \label{eq:diffEq_nofg2}
\end{equation}
Ratio between Eq. \ref{eq:diffEq_nofg1} and \ref{eq:diffEq_nofg2} yields a constant ratio $k_2$ across the thermal video, 
\begin{equation}
    k_2 = \frac{1-\epsilon_2}{1-\epsilon_1} = \frac{a_1}{a_2}\frac{I_2(t) - \tilde{I}_2(t)}{I_1(t) - \tilde{I}_1(t)}
    \label{eq:k2_const}
\end{equation}
\noindent{\bf Solution in dynamic scenes:}
The constraints described in Eq. \ref{eq:k1_const} and \ref{eq:k2_const} show that having access to robust ratios $k_1$ and $k_2$ based on locations that undergo changes in background radiation (either static or dynamic) along with a smooth signal $\tilde{I}_m(t)$, yields spectral band emissivity:
% The constraints described in Eq. \ref{eq:k1_const} and \ref{eq:k2_const} show that based on identifying locations where the background is static or dynamic, along with having access to a smooth signal $\tilde{I}_m(t)$, we can try to estimate the ratios $k_1, k_2$ in a robust fashion across as shown in Fig. \ref{}. With known $k_1$ and $k_2$ values estimating emissivity becomes straightforward, 
\begin{equation}
    \begin{aligned}
        \epsilon_1 = \frac{k_2 - 1}{k_2 - k_1}, \quad
        \epsilon_2 = k_1 \frac{k_2 - 1}{k_2 - k_1}
    \end{aligned}
    \label{eq:e1e2_from_k}
\end{equation}

\vspace{-0.2in}
\subsubsection{Optimization Framework}

Algorithm~\ref{alg:dbvt} summarizes our optimization procedure. Our goal is to jointly estimate emissivities $\epsilon_m$ and smooth signals $\tilde{I}_m(t)$ using the constraints in Eq.~\ref{eq:k1_const} and Eq.~\ref{eq:k2_const}. The recovered emissivities are then used in Eq.~\ref{eq:solve_ToTb} to compute object and background temperatures $T_o(t)$ and $T_b(t)$.

\begin{algorithm}[!t]
\caption{\small Dual-band separation (uncalibrated)}
\label{alg:short_dualband}
% \small
\footnotesize
\begin{algorithmic}[1]
\Procedure{DBVT}{$\mathbf{I_1},\mathbf{I_2},a_1,a_2,b_1,b_2$}
\State \texttt{// Initialize Variables}
\State $\tilde I_1(t), \tilde I_2(0)$, $I_m^\epsilon(t)\!\leftarrow\!0$, $\epsilon_m\!\leftarrow\!rand(0,1)$
\While{$\|\Delta \mathcal{L}\| > \delta_L$}
\State $\ddot I_m(t) \gets I_m(t) - I_m^{\epsilon}(t)$ \Comment{noise-corrected}
% \State \texttt{// Noise Corrected Estimate}
% \State $\Ddot{I}_m(t) = I_m(t) - I_m^{\varepsilon}(t)$
\State Estimate $\tilde I_2(t)$ using Eq. ~\ref{eq:i2_smooth} \Comment{Smooth Signal}
\State Estimate $\hat{I}_2(t)$ using Eq. ~\ref{eq:I2hat} \Comment{complete Signal}
\State Compute: $\mathcal{L}_{\textnormal{smooth}}, \mathcal{L}_{\textnormal{Huber}}, \mathcal{L}_{\textnormal{MSE}},\mathcal{L}_{\textnormal{noise}}$
\EndWhile
\State Recover $T_o,T_b \leftarrow$ Eq. ~\ref{eq:solve_ToTb}
\State \textbf{return} $\epsilon_m, T_o, T_b$
\EndProcedure
\end{algorithmic}
\label{alg:dbvt}
\end{algorithm}

\noindent{\bf Noise Modeling:}
Thermal videos captured with spectral filters suffer from low SNR. We explicitly model per-pixel, per-timestep noise $I_m^{\varepsilon}(t)$ and regularize it to follow a zero-mean distribution:
\begin{equation}
    \mathcal{L}_{\textnormal{noise}}^{L2} = \| I_{m}^{\varepsilon}(t) \|^2_2, \quad 
    \mathcal{L}_{\textnormal{noise}}^{M} = \frac{1}{T} \sum_{i=1}^{T} I_{m}^{\varepsilon}(i).
\end{equation}
For the rest of our optimization we use a noise-corrected signal $\Ddot{I}_m(t)$, where, $\Ddot{I}_m(t) = I_m(t) - I_m^{\varepsilon}(t)$.
% We define the noise-corrected signal as $\Ddot{I}_m(t) = I_m(t) - I_m^{\varepsilon}(t)$ and use this in the rest of our optimization.

\noindent{\bf Smooth Signal Estimation:}
We estimate smooth, emission-dominant signals $\tilde{I}_m(t)$ by leveraging the ratio constraint $k_1$. Specifically, $\tilde{I}_2(t)$ is recursively constructed from $\tilde{I}_1(t)$ as:
\begin{equation}
    \tilde{I}_2(t) = \tilde{I}_2(t-1) + k_1\frac{a_2}{a_1}(\tilde{I}_1(t) - \tilde{I}_1(t-1)).
    \label{eq:i2_smooth}
\end{equation}
We optimize for $\tilde{I}_1(t)$ and an unknown offset $\tilde{I}_2(0)$ such that $\tilde{I}_m(t)$ remains smooth while staying close to $\Ddot{I}_m(t)$. Smoothness is enforced using a second-order prior:
\begin{equation}
    \mathcal{L}_{\textnormal{smooth}} = \| \tilde{I}_m(t-1) - 2\tilde{I}_m(t) + \tilde{I}_m(t+1)\|^2.
\end{equation}
To ensure that estimated smooth signal remains consistent with the measured quantities while being robust to abrupt changes in the background, we introduce a normalized Huber loss with a threshold $\delta$ as follows:
% To ensure consistency with measurements while being robust to abrupt background variations, we use a normalized Huber loss:
\begin{equation}
    \mathcal{L}_{\textnormal{Huber}} =
    \begin{cases}
        \frac{1}{2}\|\Ddot{I}_m^n - \tilde{I}_m^n\|^2 & |\cdot| \leq \delta \\
        \delta (|\Ddot{I}_m^n - \tilde{I}_m^n| - \frac{1}{2}\delta) & \text{otherwise}.
    \end{cases}
\end{equation}

\noindent Here, superscript $n$ represents the normalization of the signals based on $\Ddot{I}_m$ by subtracting from its mean and dividing by its standard deviation. 

% \noindent{\bf Reconstruction via $k_2$:}
% Finally we regularize the estimated quantities by reconstructing the measured signal using the estimated smooth signal via the constraint $k_2$. The reconstructed signal $\hat{I}_m(t)$ can be written as:
% Finally we add a reconstruct the original measured signal from the estimated smooth signal and via the constraint on 
% To enforce the second constraint, we reconstruct $\hat{I}_m(t)$ from $\tilde{I}_m(t)$ using $k_2$:
\noindent{\bf Reconstruction via $k_2$:}
We further regularize the estimates by reconstructing the measured signal from the smooth signal using the constraint $k_2$. The reconstructed signal $\hat{I}_m(t)$ is given by:
\begin{equation}
    \hat{I}_2(t) = \tilde{I}_1(t) + k_2\frac{a_2}{a_1}(\Ddot{I}_1(t) - \tilde{I}_1(t)).
    \label{eq:I2hat}
\end{equation}
We minimize the discrepancy between reconstructed and observed signals using a mean squared error objective:
\begin{equation}
    \mathcal{L}_{\textnormal{MSE}} = \| \Ddot{I}_m^n - \hat{I}_m^n\|^2.
\end{equation}

\noindent{\bf Emissivity Optimization:}
Although emissivities can be analytically recovered from $k_1$ and $k_2$ (Eq.~\ref{eq:e1e2_from_k}), this is unstable under noise and may yield invalid solutions. Instead, we directly optimize $\epsilon_m$ values as proxies for $k_1$ and $k_2$.

% \noindent{\bf Optimization of emissivities:}
% Although emissivities can be derived from $k_1$ and $k_2$ (Eq.~\ref{eq:e1e2_from_k}), this is unstable under noise and may yield invalid solutions due to inconsistent constraint intersections (Eq.~\ref{eq:k1_const},~\ref{eq:k2_const}). We therefore directly optimize $\epsilon_m$ as proxies for $k_1$ and $k_2$.

\vspace{-0.4cm}
\paragraph{Overall Objective:}
We jointly optimize $\tilde{I}_1(t)$, $\tilde{I}_2(0)$, $\epsilon_m$, and $I_m^{\varepsilon}(t)$ (for $m \in \{1,2\}$) by minimizing:
\begin{equation}
    \underset{\tilde{I}_1(t),\ \tilde{I}_2(0),\ \epsilon_m,\ I_m^\varepsilon(t)}{\arg\min} \;
    \begin{aligned}
        &\gamma_1 \mathcal{L}_{\textnormal{smooth}} 
        + \gamma_2 \mathcal{L}_{\textnormal{Huber}} 
        + \gamma_3 \mathcal{L}_{\textnormal{MSE}} \\
        &+ \gamma_4\mathcal{L}_{\textnormal{noise}}^{L2} 
        + \gamma_5\mathcal{L}_{\textnormal{noise}}^{M}.
    \end{aligned}
    \label{eq:full_opt}
\end{equation}
Here, $\gamma_i$ denote the corresponding loss weights. This formulation balances temporal smoothness, measurement fidelity, and robustness to noise, enabling reliable recovery of emissivity and temperature in dynamic scenes.

\vspace{-0.05in}
\section{Experimental Results}\label{sec:exp}
\vspace{-0.05in}
We demonstrate our calibrated and uncalibrated approaches using both simulations and real data with time-varying emissions and reflections and compare to three baselines.

\noindent{\bf Baselines:}
We compare our method against representative approaches for reflection reduction and emission estimation from both pyrometry and computer vision literature \cite{araujo2017multi, dualWavelengthPryo, Grimming2023}, along with a naive least-squares optimization baseline.
The recent BCP method \cite{Grimming2023} assumes that the brightest pixels within a local patch approximate near-blackbody emitters, using this prior to suppress reflections. We also include a traditional dual-wavelength pyrometry baseline \cite{dualWavelengthPryo, araujo2017multi}, which neglects background radiation and models object emissivity as a gray body at the selected wavelengths. Lastly, our naive baseline jointly estimates emissivity, object temperature, and background temperature by minimizing an image reconstruction loss from Eq.~\ref{eq:multi_band_LT} in the least-squares sense. To ensure stability, we perform optimization from five random initializations and retain the result with the lowest objective value for this method.

\begin{figure}[t!]
    \centering
    \includegraphics[width=1.0\linewidth, trim={0cm 66cm 30cm 0cm}, clip]{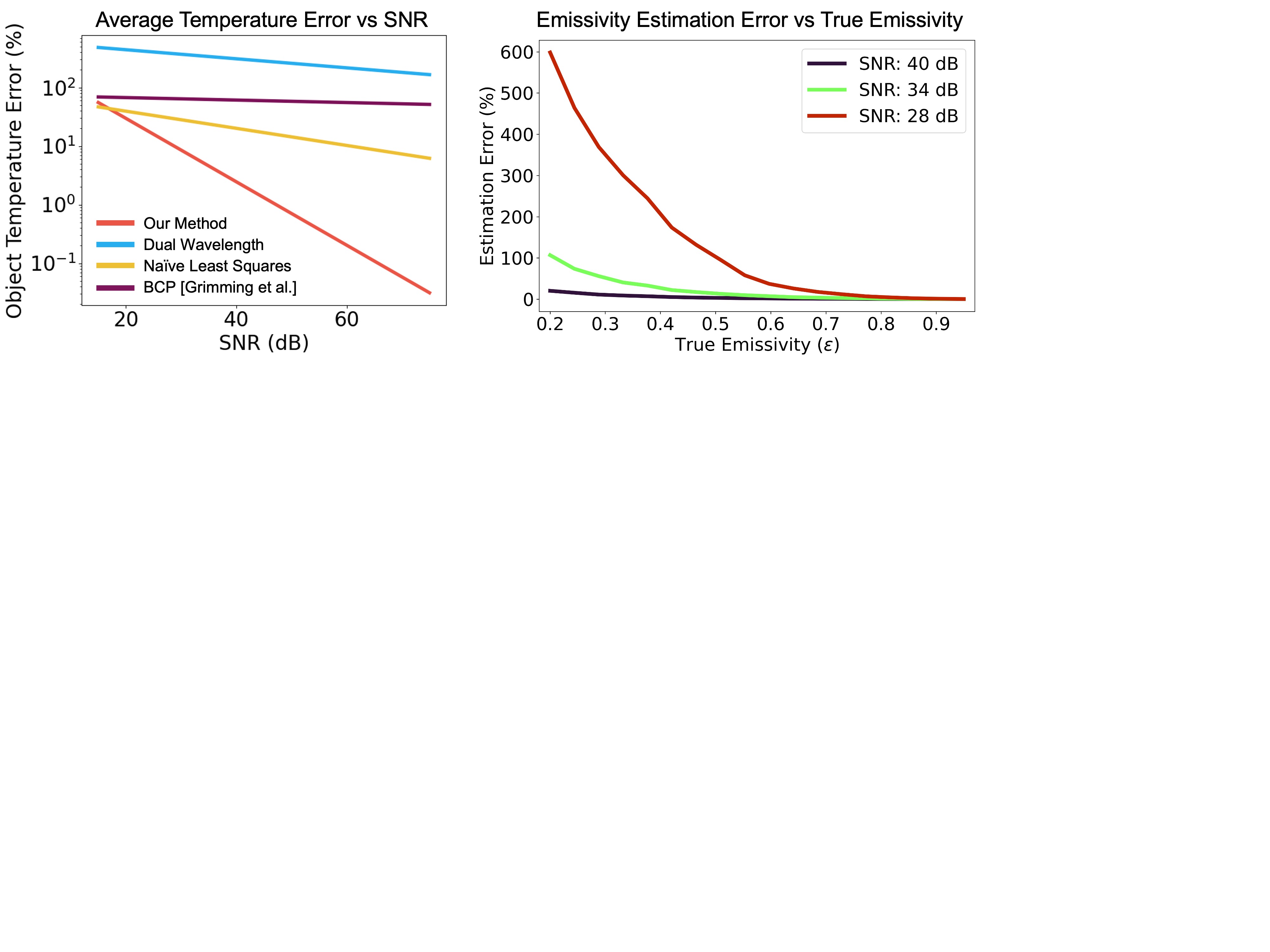}
    % \vspace{-0.5cm}
    \caption{\textbf{[Left]} Comparison of our method with a recent BCP \cite{Grimming2023} technique, naive multi-wavelength, and dual-wavelength pyrometry \cite{araujo2017multi, dualWavelengthPryo} on simulated thermal videos of different materials sourced from the spectral library \cite{ECOSTRESSv1, ECOSTRESSv2}. The naive least squares result uses the best of five initializations selected based on the one that achieved the least objective. All methods degrade at high noise; ours improves significantly at moderate noise (log scale). See our plot in the supplementary for per-material comparisons. \textbf{[Right]} Shows the error in estimated emissivities from our method for a range of emissivities under varying noise levels in simulation.
    % Comparison of our method with a recent BCP \cite{Grimming2023} technique to remove reflections, a naive multi-wavelength approach and a traditional dual-wavelength pyrometry technique \cite{araujo2017multi, dualWavelengthPryo} using simulated thermal videos of different materials sourced spectral library. The naive least squares result uses the best of five initializations (oracle). Our method performs better at moderate noise and plot is shown in log scale.
    % Comparison of our method with a recent BCP \cite{Grimming2023} technique to remove reflections, a naive multi-wavelength approach and a traditional dual-wavelength pyrometry technique \cite{araujo2017multi, dualWavelengthPryo} using simulated thermal videos of different materials sourced from spectral library \cite{ECOSTRESSv1, ECOSTRESSv2}. For naive least squares, we run the optimization with five initializations and select one that achieved the least objective compared to ground truth (which we will not have access to at test time). At high noise levels, all methods have a large error as the problem is too under constrained. As noise decreases to more reasonable levels, our method performs significantly better. Note the log scale on the plots. \textbf{[Right]} Emissivity estimation error for a range of emissivities under varying noise levels.
    }
    \label{fig:snr_error}
    % \vspace{-\intextsep}
    \vspace{-0.1in}
\end{figure}

\noindent{\bf Experimental Setup:}
Our setup uses two FLIR Boson thermal cameras-a performance-grade unit ($\leq 40$ mK NETD) and an industrial-grade unit ($\leq 20$ mK NETD)-operating at 640$\times$512 resolution with a $24^\circ$ HFOV lens. The cameras are placed side by side to minimize parallax; we opt against using a beam splitter due to its significant SNR penalty. We capture data using spectral filters centered at 8.5~$\mu$m, 9.5~$\mu$m, 10.6~$\mu$m, and 12.1~$\mu$m with FWHM values of 0.5~$\mu$m, 0.5~$\mu$m, 1.5~$\mu$m, and 0.5~$\mu$m, respectively. The filters are mounted on an FW103H/M motorized wheel, with the industrial-grade camera positioned in front of the wheel. Ground-truth surface temperatures are recorded using a TC-08 data logger with Type-T thermocouples.

\subsection{Simulations}
\begin{figure}[t!]
    \centering
    % \includegraphics[width=1.0\linewidth, trim={0cm 82.9cm 0cm 0cm}, clip]{pics/spatial_error_rebut4.jpg}
% \includesvg[inkscapelatex=false, width=1.00\linewidth]{pics/spatial-error-svg}
    \includegraphics[width=1.0\linewidth]{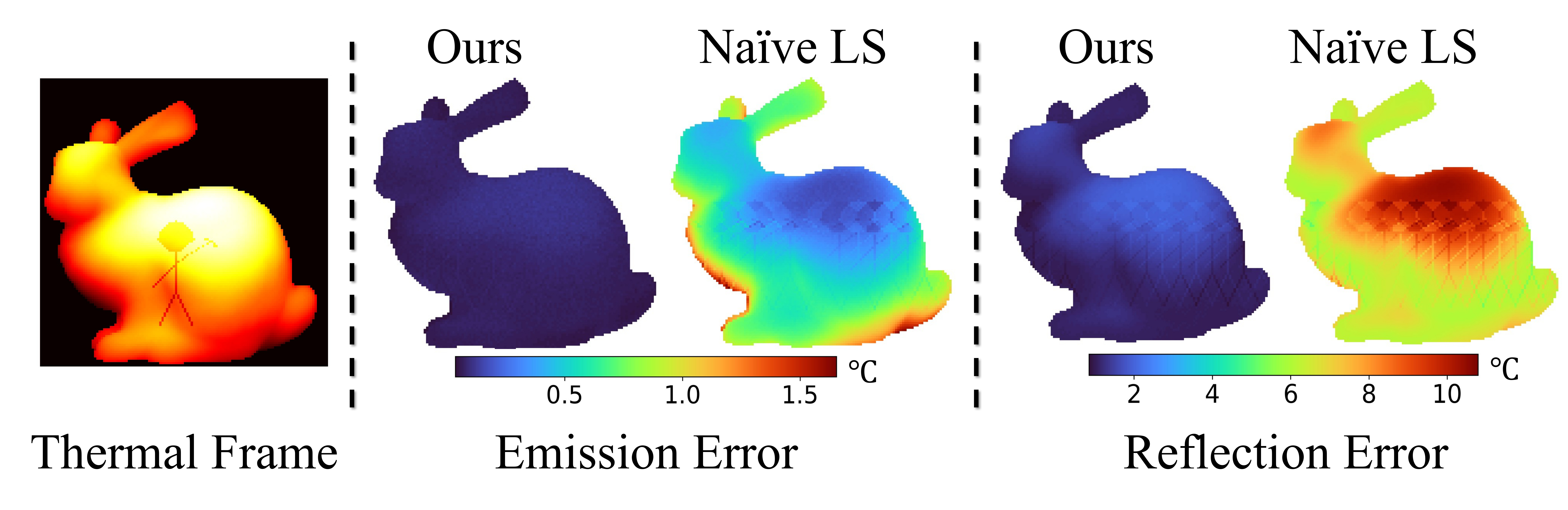}
    \caption{Spatial temperature errors (in $^\circ C$) averaged over video frames for our method and naive least squares (LS). The thermal video (left) was obtained through a combination of heat simulation and a moving toy background based on Eq. \ref{eq:multi_band_LT} with added noise.}
    %\vspace{-0.15in}
    \label{fig:spatial-error2}
\end{figure}

We evaluate our method using controlled heat-transfer simulations~\cite{Narayanan2024Shape} on objects (e.g., Stanford Bunny) against a dynamically varying background. Thermal sequences are generated via Eq.~\ref{eq:multi_band_LT} over a wide range of emissivities sampled from spectral libraries~\cite{ECOSTRESSv1, ECOSTRESSv2} (see supplementary for details). 
As shown in Fig.~\ref{fig:snr_error} (left), our method consistently outperforms all baselines in recovering object temperature across noise levels. Spatial error maps in Fig.~\ref{fig:spatial-error2} show accurate reconstruction, with minor errors near object boundaries due to reduced thermal contrast from reflected background. 
Fig.~\ref{fig:snr_error} (right) demonstrates that our uncalibrated approach can also recover emissivity across noise levels, performing well at moderate-to-high emissivities, while low-emissivity and high-noise regimes remain challenging.

\begin{table}[!t]
    \centering
    \resizebox{\linewidth}{!}{
        \begin{tabular}{c|c|c|c|c}
            \toprule
             Loss-terms & Reconstruction term & Smoothing term & Huber loss & Noise terms\\
             \midrule
             $\%$ reduction & 90.10\% & 56.73\% & 11.31\% & 11.64\%\\
             \bottomrule
        \end{tabular}
    }
    \caption{Percentage error reduction in temperature estimates by adding the loss terms individually (while including the other three loss terms), averaged over a sweep of SNR levels in simulation.}
    \label{tab:loss_ablation}
    % \vspace{-\intextsep}
\end{table}

\noindent{\bf Ablation of Loss Terms:} Table~\ref{tab:loss_ablation} quantifies the contribution of each loss term by measuring performance gains when added to the remaining losses. Reconstruction loss is the most critical component, followed by smoothing and Huber losses, while the noise term becomes beneficial under high-noise conditions.
% \noindent{\bf Ablation of Loss Terms:} Table ~\ref{tab:loss_ablation} shows the performance improvement obtained by adding each loss term compared to an optimization with other loss terms. Reconstruction loss is obviously the most important term followed by smoothing and Huber loss. The noise term is useful at high noise levels.

% We individually disable each loss term in the objective (\ref{eq:full_opt}). Reconstruction loss is obviously the most important term followed by smoothing, huber and noise optimization. The noise term is useful at high noise levels. \emph{See plot in Supplementary.} 

\noindent{\bf Filter Selection:} Spectral filters are selected based on two factors: SNR of the thermal image captured and the change in emissivity across the filters. For the first, we choose the image without a filter (8-14$\mu m$) since the SNR is highest. Then, we choose the second filter, $9.5\mu m$, that has the highest condition number of the resulting emissivity matrix $\mathbf{E}$ from Eq. \ref{eq:img_form_matform_vfgbg} using emissivities sourced from the spectral library \cite{ECOSTRESSv1}. \emph{See full ablation in Supplementary.} 

%\input{figures/condition_number}
%\paragraph{Choice of Spectral Filters:} The selection of spectral filters for thermography plays a crucial role in distinguishing emitted and reflected light. Figure \ref{fig:condition-number} illustrates the condition number of the matrix $\mathbf{E}$ from Eq. \ref{eq:img_form_matform_vfgbg} for different spectral filter choices in our experiments, using emissivities sourced from the spectral library \cite{ECOSTRESSv1}. The choice of spectral bands involves a trade-off between the condition number of the resulting emissivity matrix and the noise introduced by each filter. Generally, wider spectral bands reduce noise levels, so we select 8--14$\mu m$ as one of the spectral bands. The 9.5$\mu m$ filter achieves the lowest condition number, which we select as the second band.

% \vspace{-0.05in}
\subsection{Real Experiments}

\begin{figure}[!t]
    \centering
   \includegraphics[width=1.0\linewidth, trim={0 10cm 0 0.0cm}, clip]{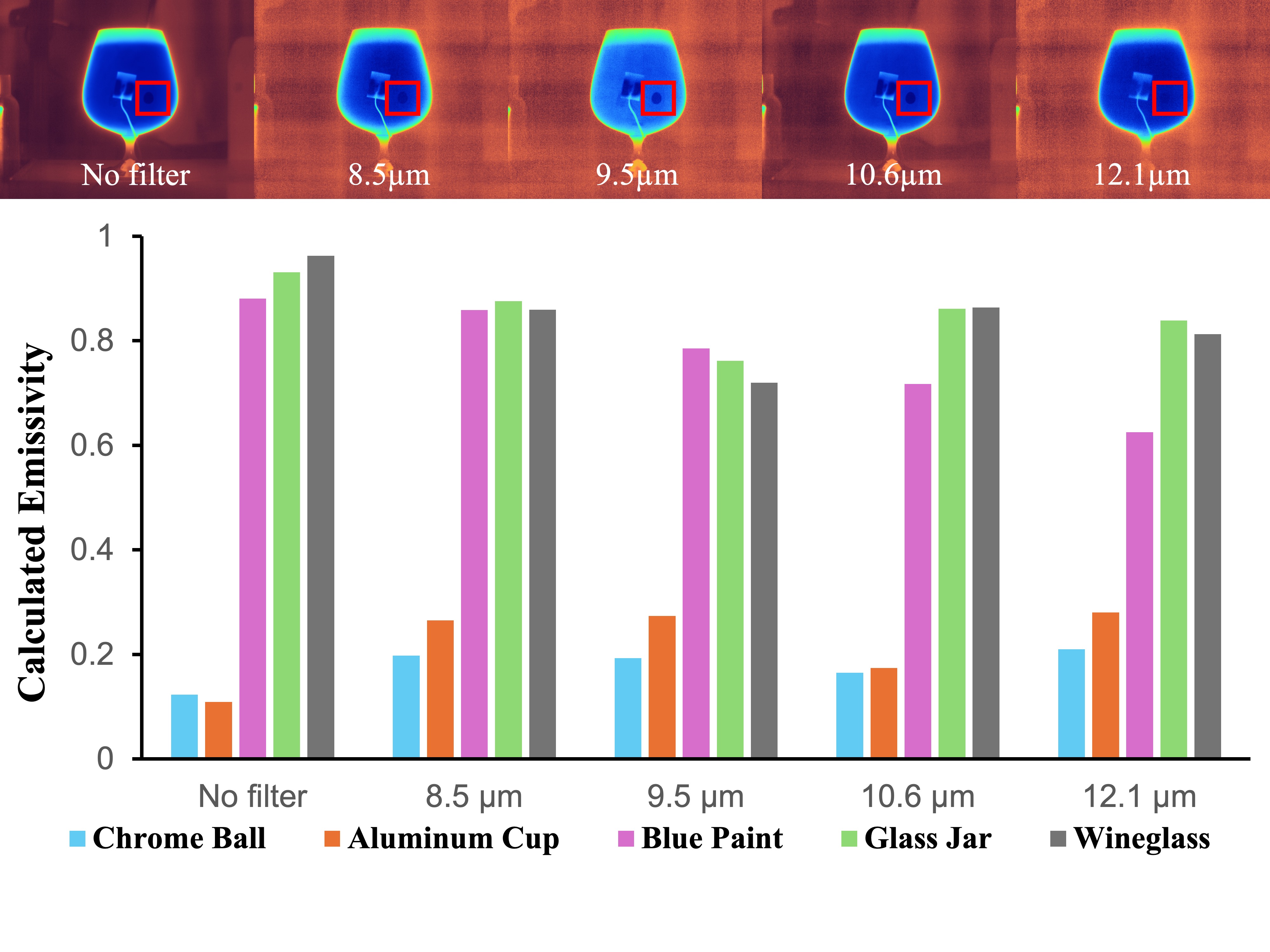}
   % \includegraphics[width=0.9\linewidth]{pics/calibration-wine-glass.png}\\
      %\includegraphics[width=1.00\linewidth]{pics/calibration-test.jpg.png}\\
      % \includesvg[inkscapelatex=false, width=1.00\linewidth]{pics/emissivity-plot.svg}\\
      %\includegraphics[width=1.00\linewidth]{pics/plot-pdf.pdf}
      % \vspace{-0.05in}
    \caption{% Images on the bottom show an example figures of different filters used in the calibration procedure, with a red bounding box marking the reflected blackbody on the object. A thermocouple is attached next to this black body reflection to measure the object's temperature. The above plot shows the calibrated emissivity values for different objects and materials using the proposed calibration technique. While an exact comparison of this technique with the spectral library is not practical, our emissivities lie in ballpark with the numbers reported in spectral library \cite{ECOSTRESSv1}. The reported emissivity values at 8$\mu m$-14$\mu m$ for aluminum metal and plate glass are 0.05 and 0.87, our method achieves an emissivity of 0.1 for the aluminum cup and 0.93 for the glass jar indicating close proximity with similar materials.
    {\bf [Top]} Images captured with different spectral filters used for calibration. The red bounding box highlights the reflected blackbody on the object. A thermocouple is placed next to this reflection to measure the object's temperature. {\bf [Bottom]} Calibrated emissivity values for various materials using our technique align the closely with reported values \cite{design1st_emissivity_values, ECOSTRESSv2} as shown in Tab.~\ref{tab:emiss_comp}. 
    % Specifically, for the 8–14$\mu m$ range, the spectral library lists emissivities of 0.05 for aluminum and 0.87 for plate glass, while our method estimates 0.1 for an aluminum cup and 0.93 for a glass jar, demonstrating strong agreement with similar materials.
    }
    \label{fig:emiss-calib}
    \vspace{-0.1in}
\end{figure}

\noindent{\bf Emissivity Calibration and Comparison:} Figure \ref{fig:emiss-calib} illustrates our emissivity calibration setup. A thermocouple is placed near a specular background reflection to measure the object's temperature. To obtain a robust estimate of the object's emissivity using Eq. \ref{eq:emiss_calib_eq}, we capture images of the object and its background reflections at different temperatures. This is achieved by heating the object with hot water or a hot air gun and varying the reflected blackbody temperature. The resulting spectral-band emissivities for multiple objects are shown in Figure~\ref{fig:emiss-calib}. As summarized in Table~\ref{tab:emiss_comp}, our estimates closely match the reference values in \cite{design1st_emissivity_values, ECOSTRESSv2}. Our method also outperforms the FLIR Systems reflector method \cite{flir_thermography_reference_2019} for low-emissivity surfaces, because our method leverages specular reflections and is less sensitive to inaccuracies compared to the reflector approach.

\begin{table}
    \centering
    \resizebox{\linewidth}{!}{
        \begin{tabular}{c|c|c|c|c|c}
            \toprule
             Methods & Chrome Ball & Al. Cup & Blue Paint & Glass Jar & Wineglass\\
             \midrule
             Reference Emissivity & 0.1 & 0.05 & 0.87 & 0.95 & 0.95 \\
             \midrule
             Reflector Method \cite{flir_thermography_reference_2019} & 0.43 & 0.16 & 0.88 & 0.93 & 0.97\\
             Ours & \textbf{0.12} & \textbf{0.10} & \textbf{0.88} & \textbf{0.93} & \textbf{0.96} \\
             \bottomrule
        \end{tabular}
    }
    \caption{Comparison of emissivity estimates obtained using our calibration technique and the reflector method \cite{flir_thermography_reference_2019} for various materials using 8–14$\mu m$ range. Our estimates closely match those from the reflector method and show improved accuracy for low-emissivity materials. Reference emissivity values are from \cite{design1st_emissivity_values, ECOSTRESSv2} and may not reflect the exact emissivities of the tested objects.}
    \label{tab:emiss_comp}
    \vspace{-0.15in}
\end{table}

\begin{figure}[t]
    \centering
   % \includegraphics[width=1.0\linewidth, trim={0 3.0cm 11cm 0}, clip]{pics/teaser/teaser-v1.jpg}
   %\includegraphics[width=1.02\linewidth]{pics/palm-reflection-print.png}
   % \includesvg[inkscapelatex=false, width=1.00\linewidth]{pics/palm-print-reflection.svg}
   \includegraphics[width=1.0\linewidth]{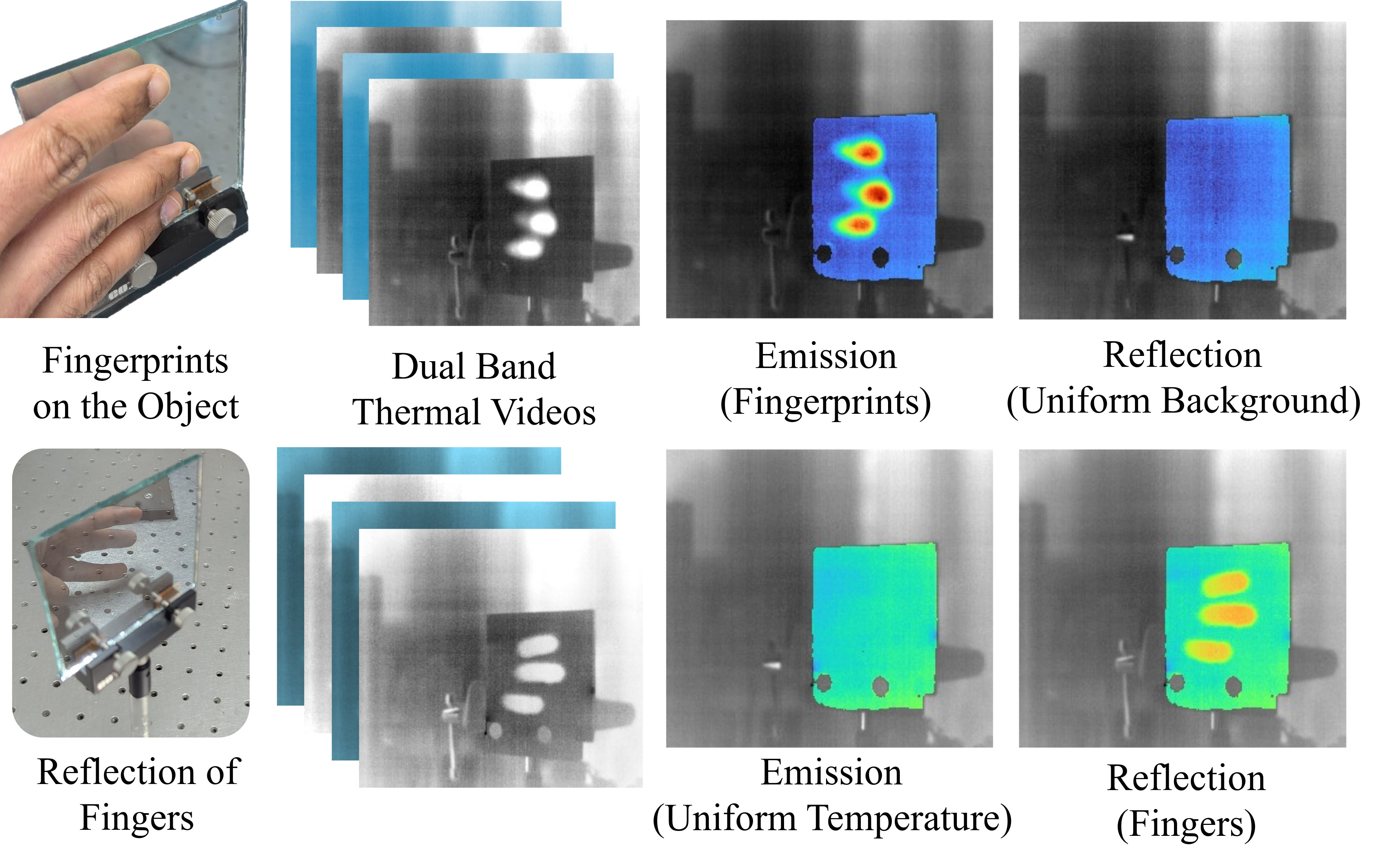}
   \vspace{-0.2in}
    \caption{Finger print versus Finger reflection: [Top] Separating finger prints on a glass plate that emit light (heat transport) from the reflection of the uniform background (light transport). [Bottom] Separating the reflection of the fingers (light transport) by the glass plate at constant room temperature (heat transport). 
    %{\bf Please see supplementary video.}
    }
    \label{fig:teaser-fig}
    \vspace{-0.15in}
    %\vspace{-\intextsep}
\end{figure}
%\vspace{-0.4cm}
%\paragraph{Reflection-Emission Separation Results:} We show results for several real-world scenes using dual thermal bands—one captured without a filter and the other with a $9.5\mu m$ central wavelength filter. The $9.5\mu m$ filter was chosen since it maximizes the emissivity differences (and the condition number of the linear system) for these objects.

\begin{figure*}[!t]
    \centering
    \includegraphics[width=1.0\textwidth, trim={0cm 21.0cm 2.8cm 0}, clip]{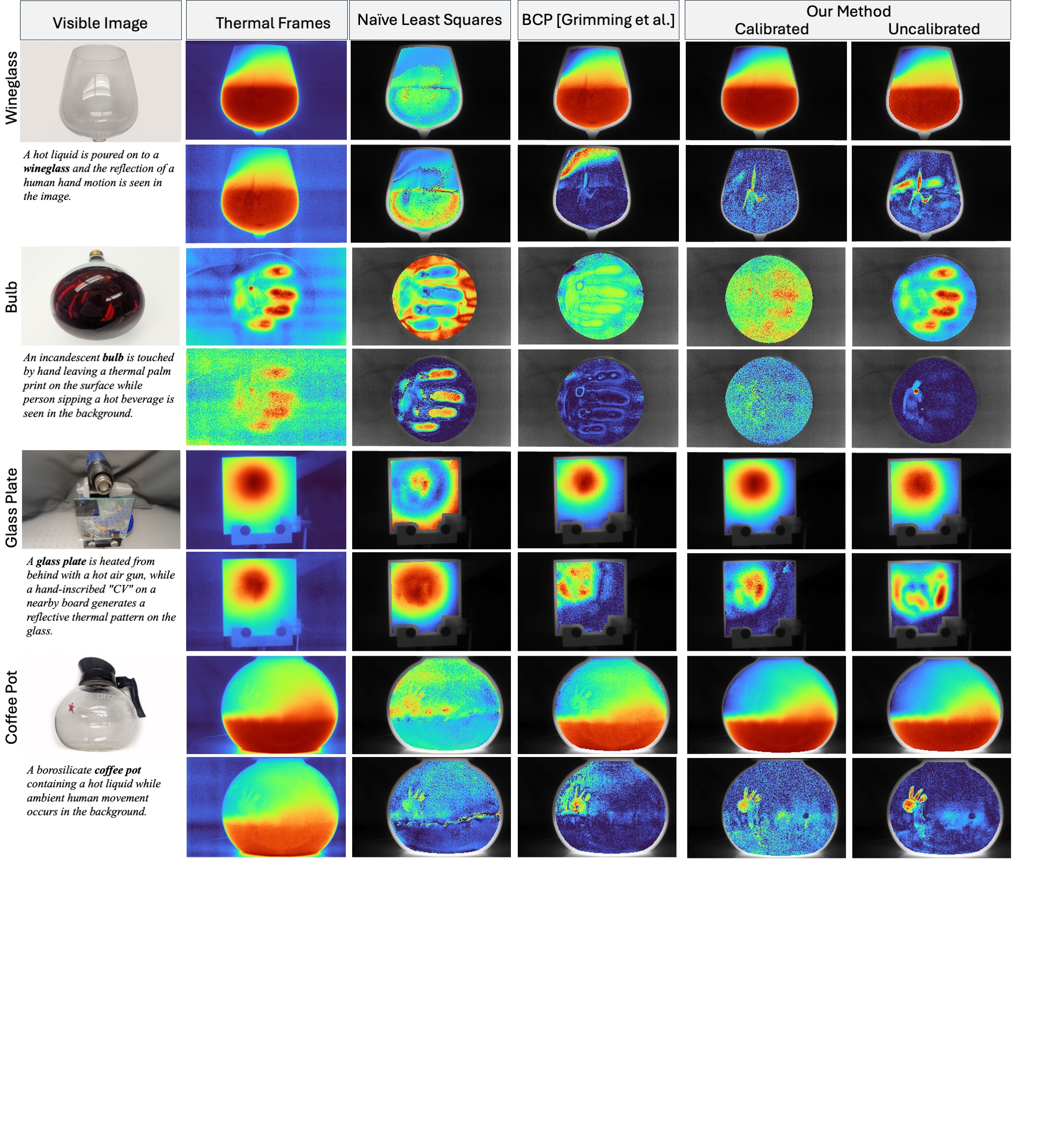}
    \caption{ 
    Reflection--emission separation results for objects heated using different methods (hot liquid, hand contact, and hot air). For each object, the first row shows the estimated emission, and the second row shows the mean-subtracted reflection. Input thermal frames from two spectral bands are shown on the left. Since per-pixel temperature ground truth is unobtainable, we record temperatures at sparse locations using a thermocouple. The wineglass and coffee pot reached peak temperatures of $63.6^\circ C$ and $63.1^\circ C$, respectively. The corresponding estimation errors were $1.72\%$, $5.04\%$, $14.6\%$, and $31.68\%$ for the uncalibrated, calibrated, BCP, and naive approaches on the wineglass video, and $5.34\%$, $0.36\%$, $6.62\%$, and $45.5\%$ for the coffee pot video. Our method cleanly separates emission and reflection even in challenging cases, such as isolating the ``CV'' reflection on glass or distinguishing a palm print on a bulb from a person sipping coffee. 
    % {\bf Please see supplementary video for best visualizations.}
    % Reflection-emission separation results for various objects heated through different methods, including hot liquid, hand contact, and a hot air gun. For each object, the first row shows the emission frame, while the second row presents the mean-subtracted reflection image. On the left, we display two input thermal frames from dual spectral bands. Note that per-pixel temperature ground truth is practically unattainable so we captured temperature using a thermocouple at a few points. The wineglass and coffee pot reached maximum temperatures of \( 63.6^\circ C \) and \( 63.1^\circ C \), respectively. The percentage error of estimations compared to these measurements was \( 1.72\% \), \( 5.04\% \), \( 14.6\% \) and \( 31.68\% \) for uncalibrated, calibrated, BCP and naive optimization approaches in the wineglass video, and \( 5.34\% \), \( 0.36\% \), \( 6.62\% \), and \( 45.5\% \) for the coffee pot video. Visually, our method effectively separates emission and reflection videos, even in challenging scenarios, such as isolating the reflection of ``CV'' on a glass plate or distinguishing a palm print on a bulb from a person sipping coffee.
    }
    \label{fig:results_img}
    \vspace{-\intextsep}
    \vspace{0.3cm}
\end{figure*}

%\vspace{-0.4cm}
\noindent{\bf Reflection-Emission Separation Results:} 
%We show visual results for several real-world scenes. 
%We show results for several real-world scenes using dual thermal bands—one captured without a filter and the other with a $9.5\mu m$ central wavelength filter. The $9.5\mu m$ filter was chosen since it maximizes the emissivity differences (and the condition number of the linear system) for these objects.
First, we illustrate the difference between reflection due to light transport and emission due to heat transport with the same object: the reflection of fingers moving in front of a glass plate vs. the finger prints dissipating on the glass plate. Our method separates these components well, as shown in Figure \ref{fig:teaser-fig}. We demonstrate separation for several interesting and challenging scenes whose descriptions are given in Figure~\ref{fig:results_img}. 
Our results demonstrate the effectiveness of our optimization in isolating background reflections from object emissions, even in visually imperceptible cases such as the glass plate, while simultaneously providing meaningful temperature estimates validated via contact thermometry.
While our calibrated approach tends to be noisier in low-signal regions due to its analytical nature and the absence of a temperature prior, it achieves decent separation results for high-emissivity objects with good signal, as shown in Figure~\ref{fig:results_img}. The estimated emissivities from our uncalibrated approach deviate by an average of $0.17, 0.15, 0.19$, and $0.15$ from our calibrated method across the dual spectral bands. Also, our uncalibrated approach has an average temperature difference of $1.72\%$ and $5.34 \%$ with wineglass and coffee pot videos compared to contact-measurements with a thermocouple, highlighting the ability of our method to perform thermography in complex, dynamic real-world scenarios for the first time. Best viewed in the Supplementary video.
\section{Limitations}
Thermographic separation of reflection and emission is a highly ill-posed problem, with no widely accepted standard for analysis in general scenarios. While our work makes significant progress in resolving ambiguities between emitted and reflected radiation, several challenges remain: (i) We assume that changes in background radiation are uncorrelated with the object signal. While this holds for reflections caused by object motion, it may not be valid in scenarios such as a room warming uniformly. (ii) Low-cost microbolometers with spectral filters have limited sensitivity, making it difficult to detect small temperature differences in low-emissivity objects due to poor signal-to-noise ratio. 
% In such scenarios, learned inductive priors based of heat flow \cite{Dashpute_2023_CVPR, bao2023heat} 

\vspace{-0.05in}
\section{Conclusion}
Heat and light transport jointly encode the material and geometric properties of a scene~\cite{Narayanan2024Shape,Dashpute_2023_CVPR,TanakaFar}. Disentangling these components is essential for robust thermal interpretation in unconstrained environments. We presented a dual-band approach with both calibrated and uncalibrated formulations, supported by a calibration procedure that estimates spectral-band emissivities from measured temperatures. We further introduced a theory and optimization strategy that leverages variations in emissivity and temperature to add structure to an otherwise ill-posed decomposition. As thermal cameras become increasingly accessible, exploring the interaction between light and heat transport remains a promising direction for advancing thermal imaging and its applications.

\paragraph{\textbf{Acknowledgements:}} This work was partly supported by NSF grants IIS-2107236, and NSF-NIFA AI Institute for Resilient Agriculture.

{
    \small
    \bibliographystyle{ieeenat_fullname}
    \bibliography{main}
}

\appendix
\clearpage
\section{Calibration and Pre-Processing}
In this section, we provide more details on our camera calibration procedure.

\begin{figure}[t]
    \centering
    \includegraphics[width=1.0\linewidth, trim={0 0cm 0cm 0}, clip]{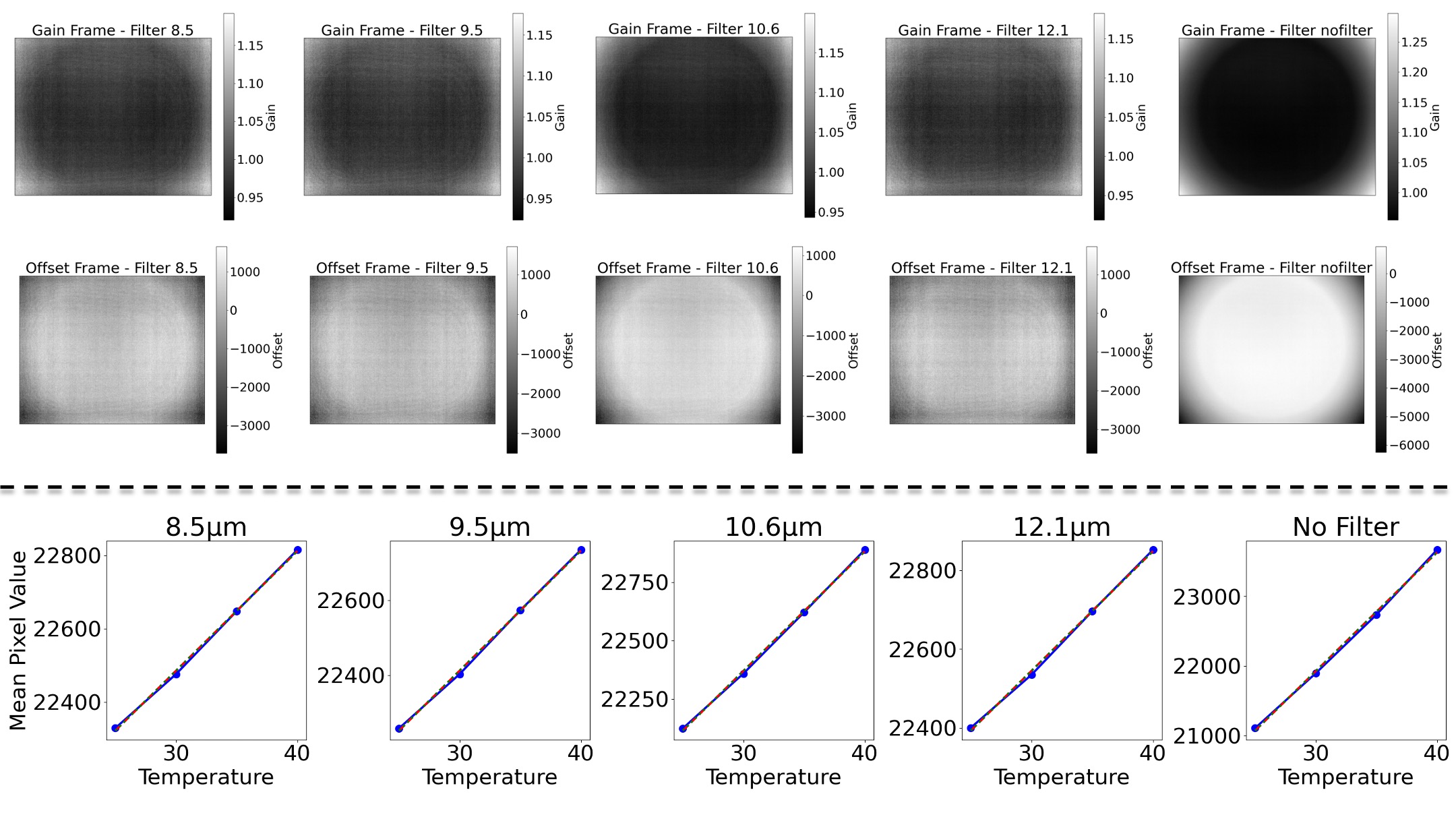}
    \caption{
    % Top: Shows the gain and offset correction images for the various spectral bands. Unlike \cite{SFTR} which only performs offset correction, we perform both gain and offset correction for accurate recovery of incoming radiation. Bottom: Shows blackbodies at different temperatures and their corresponding pixel values in {\color{blue} blue}, with the exponential Sakumo-Hattori fit\cite{sakuma1982establishing} and linear fit for the pixel values is shown in dotted {\color{red} red} and {\color{green} green} respectively. As seen, both temperature to camera counts curve looks linear and the difference between the Sakumo-Hattori and linear curve is less than a couple of counts, which is much below the noise floor of the thermal camera.
    Top: Shows the gain and offset correction images for the various spectral bands. Unlike \cite{SFTR} which only performs offset correction, we perform both gain and offset correction for accurate recovery of incoming radiation. Bottom: Shows blackbodies at different temperatures and their corresponding pixel values in {\color{blue} blue}, with the exponential Sakuma-Hattori fit \cite{sakuma1982establishing} and linear fit for the pixel values is shown in dotted {\color{red} red} and {\color{green} green} respectively. As seen, both temperature to camera counts curve looks linear and the difference between the Sakuma-Hattori and linear curve is less than a couple of counts, which is much below the noise floor of the thermal camera.
    }
    \label{fig:gainOffset-fig}
    \vspace{-\intextsep}
\end{figure}
\vspace{-0.2cm}
\paragraph{Gain-Offset Correction:} 
We begin by applying gain and offset correction using the gain and offset frames shown in Figure \ref{fig:gainOffset-fig}. To calibrate, we position a blackbody at various temperatures in front of the camera, ensuring it fills the entire field of view. The true pixel value is determined as the mean pixel count across the entire image. For each pixel, we compute gain and offset values so that the corrected output matches this mean. This calibration is performed in a least-squares sense across multiple blackbody temperatures.
% We first perform gain and offset correction based on the gain and offset frames shown in Figure \ref{fig:gainOffset-fig}. To perform gain and offset calibration we place a blackbody at different temperatures in front of the camera covering its entire field of view. We calculate the mean pixel count across the entire image as the true pixel value and find a gain and offset for each pixel such that resulting gain-offset corrected value is its mean. We do this in a least squares sense for blackbodies at different temperatures.

% \vspace{-0.2cm}
% \paragraph{Data Preprocessing \& Calibration:} 
% We first correct for the narcissus effect using images of a blackbody at two known temperatures to compute both gain and offset values, compared to offset only correction used in ~\cite{SFTR}. Afterward, we calibrate the camera response function $U(T)$ using blackbody measurements at multiple known temperatures. Further details on data preprocessing and calibration are in the supplementary material.

\paragraph{Thermal Camera Calibration:} Similar to gain-offset correction, we use blackbody videos at different temperatures to perform radiometric calibration of thermal cameras. Figure \ref{fig:gainOffset-fig} presents the calibration results across various spectral bands, comparing a Planckian Sakuma-Hattori fit \cite{sakuma1982establishing} with a linear fit. As observed, the camera response is nearly linear for objects at ambient temperatures.

\section{More Experimental Results}
% In this section, we perform additional ablation studies of our proposed optimization pipeline and analyze the ability of dual-band filters to solve the proposed image formation model based on the choice of spectral filters used. Additionally, we show the breakdown of our simulation results in the camaera

\begin{figure}[t]
    \centering
    \includegraphics[width=0.75\linewidth, trim={0 18cm 40cm 0}, clip]{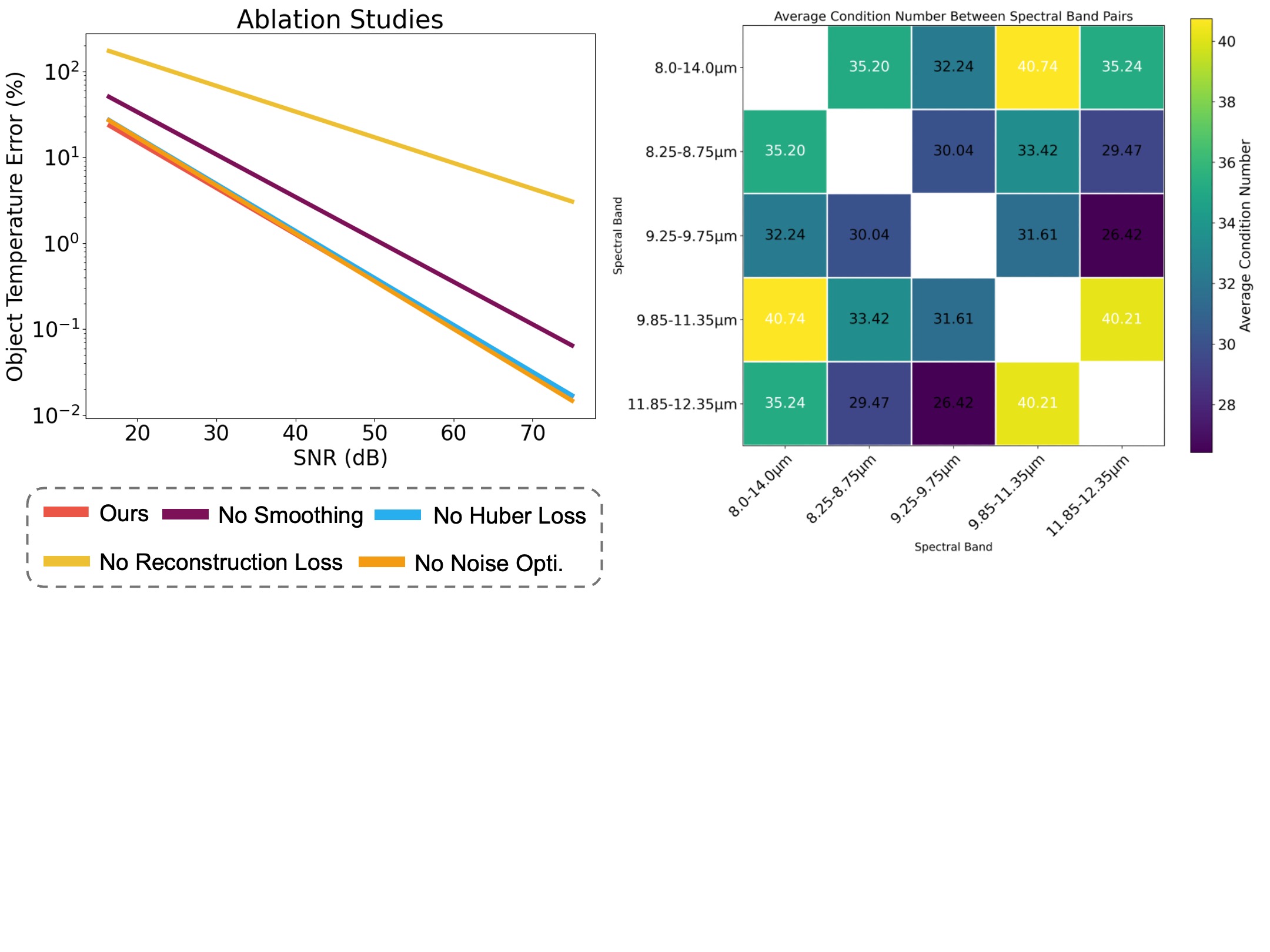}
    \vspace{-0.2in}
    \caption{ 
    Ablating the loss terms in simulation: We individually disable each loss term from the full optimization pipeline. Note the log scale on the plots. Reconstruction loss is obviously the most important term followed by smoothing, huber and noise optimization. The noise term is useful at high noise levels.
    }
    \label{fig:ablation}
    \vspace{-0.05in}
    %\vspace{-\intextsep}
\end{figure}

\paragraph{Ablation Studies:} Figure \ref{fig:ablation} presents the ablation results of our optimization, where each loss term is individually removed from the pipeline. As shown, the reconstruction loss has the most significant impact, while the smoothing and Huber losses have a comparatively smaller effect on overall accuracy. Initially, optimizing noise parameters does not provide a noticeable improvement, but at lower SNR values, failing to account for noise leads to higher errors. Fig. ~\ref{fig:ablation} represents the SNR level breakdown of Tab.~\ref{tab:loss_ablation} shown in the main paper.

\begin{figure}[t]
    \centering
    \includegraphics[width=1.0\linewidth, trim={0cm 16cm 30cm 0cm}, clip]{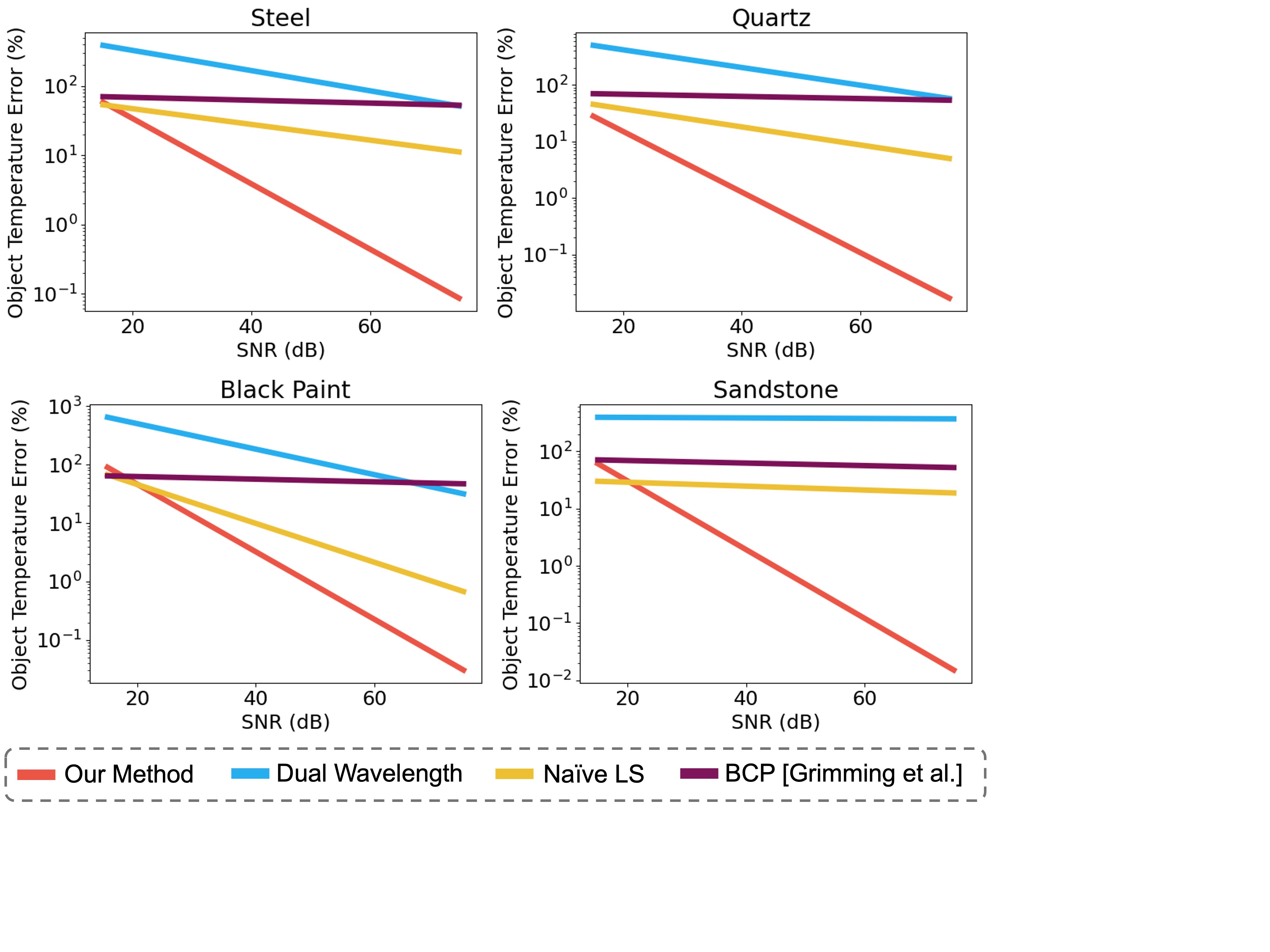}
    % \vspace{-0.5cm}
    \caption{Comparison of our method with a recent BCP \cite{Grimming2023} technique to remove reflections, a naive multi-wavelength approach and a traditional dual-wavelength pyrometry technique \cite{araujo2017multi, dualWavelengthPryo} using simulated thermal videos of different materials sourced from spectral library \cite{ECOSTRESSv1, ECOSTRESSv2}. For naive least squares, we run the optimization with five initializations and select one that achieved the least objective compared to ground truth (which we will not have access to at test time). At high noise levels, all methods have a large error as the problem is too under constrained. As noise decreases to more reasonable levels, our method performs significantly better. Note the log scale on the plots.
    }
    \label{fig:snr_error_full}
\end{figure}

\paragraph{Breakdown of Simulation Evaluation:} Fig.~\ref{fig:snr_error_full} provides a detailed breakdown of the simulation results summarized in Fig.~\ref{fig:snr_error} in the main paper.

\begin{figure}[ht]
    \centering
    \includegraphics[width=1.0\linewidth, trim={0 0cm 0cm 0}, clip]{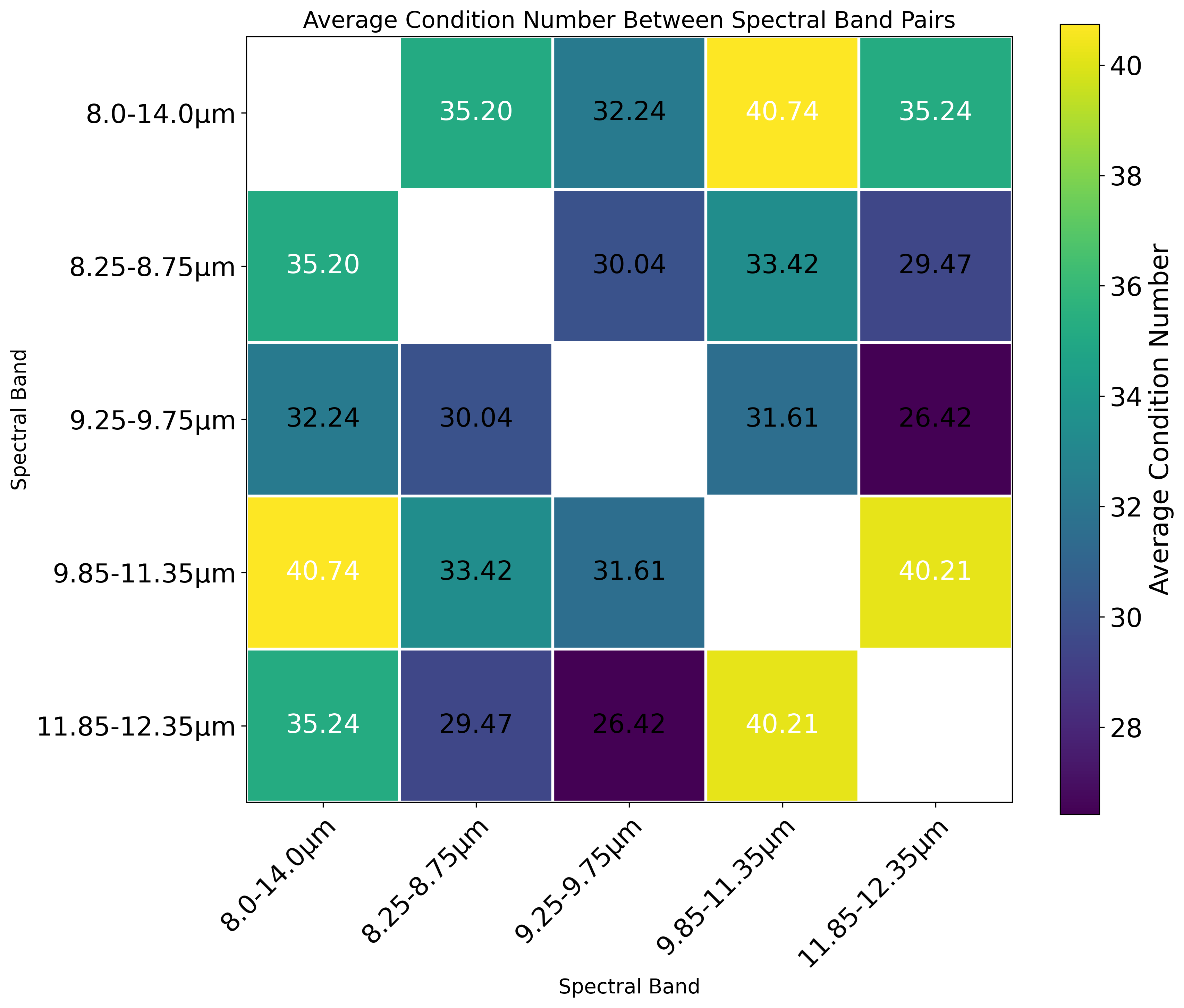}
    \caption{
    Condition numbers of emissivity matrix $\mathbf{E}$ in Eq. \ref{eq:img_form_matform_vfgbg} for different spectral band pairs, averaged over materials in the spectral library \cite{ECOSTRESSv1, ECOSTRESSv2}. 
    }
    \label{fig:condition-number}
    \vspace{-\intextsep}
\end{figure}
\paragraph{Choice of Spectral Filters:} The selection of spectral filters for thermography plays a crucial role in distinguishing emitted and reflected light. Figure \ref{fig:condition-number} illustrates the condition number of the matrix $\mathbf{E}$ from Eq. \ref{eq:img_form_matform_vfgbg} for different spectral filter choices in our experiments, using emissivities sourced from the spectral library \cite{ECOSTRESSv1}. The choice of spectral bands involves a trade-off between the condition number of the resulting emissivity matrix and the noise introduced by each filter. Generally, wider spectral bands reduce noise levels; hence, we select 8--14$\mu m$ as one of the spectral bands. Additionally, Figure \ref{fig:condition-number} shows that the 9.5$\mu m$ central wavelength filter achieves the lowest condition number, indicating a better-conditioned linear system for our application.

% \section{Implementation Details}

% \input{figures/snr_error}

\end{document}